\documentclass{TEAI}
\pdfoutput=1  
\usepackage{algorithm}     
\usepackage{algorithmic}   
\usepackage{enumitem}
\usepackage{flafter}       
\graphicspath{{figs/}}

\newcommand{\deff}{d_{\mathrm{eff}}}        
\newcommand{\dperp}{d_{\perp}}              
\newcommand{\dans}{d_{\mathrm{ans}}}        
\newcommand{\Wmat}{\mathbf{W}}              
\newcommand{\Wc}{\mathbf{W}_{C}}            
\newcommand{\dens}{\mathrm{dens}}

\newcommand{\nrollouts}{N}
\newcommand{\bbudget}{b}

\newcommand{\Ncohorts}{3{,}105}   
\newcommand{\Naudit}{3{,}096}     
\newcommand{\Ncomplete}{2{,}999}  
\newcommand{\Nframes}{49{,}680}   
\newcommand{\Nident}{49{,}549}    
\newcommand{\Ncvframes}{47{,}984} 
\newcommand{\Ngraphs}{52{,}785}   
\newcommand{\Nsteps}{28{,}622}    
\newcommand{\Nunanimous}{1{,}193} 
\newcommand{\NdefinedAP}{1{,}912} 
\newcommand{\NAPaudit}{1{,}906}   

\newcommand{\takeaway}[2]{%
  \par\addvspace{3pt}\noindent
  \fcolorbox{black!45}{black!4}{%
    \parbox{\dimexpr\columnwidth-2\fboxsep-2\fboxrule\relax}{%
      \small\textbf{Takeaway~#1.}~\emph{#2}}}%
  \par\addvspace{3pt}}

\theoremstyle{definition}

\title{From Concentration to Differentiation and Back:\\
Routing Effective Rank in MoE Reasoning Cohorts}

\author[1]{Kang Chen}
\author[1]{Sihan Zhao}
\author[1,2\dagger]{Yixin Cao}
\author[1]{Yu-Gang Jiang}
\affiliation[1]{Fudan University}
\affiliation[2]{Shanghai Innovation Institute}
\checkdata[Email]{\email{kchen24@m.fudan.edu.cn}, \email{yxcao@fudan.edu.cn}$^{\dagger}$}
\checkdata[Project]{\url{https://cckfdu.com/deff/}}

\hypersetup{
  pdftitle={From Concentration to Differentiation and Back: Routing Effective
            Rank in MoE Reasoning Cohorts},
  pdfauthor={Kang Chen, Sihan Zhao, Yixin Cao, Yu-Gang Jiang},
  pdfsubject={Mixture-of-Experts routing, test-time scaling, effective rank},
  pdfkeywords={mixture of experts, expert routing, test-time scaling,
               effective rank, reasoning cohorts}
}

\abstract{
Test-time scaling produces cohorts of reasoning rollouts, yet there is no
standard label-free account of how their internal computation reorganizes as
inference unfolds. We introduce \emph{routing effective rank} $\deff$, the
entropy-effective dimensionality of a cross-rollout graph built from MoE
expert-routing similarity. Across ten MoE configurations and five math/science
benchmarks, $\deff$ exhibits a reproducible low--high--low trajectory, with a
prominent interior maximum in $98.5\%$ of $\Ncohorts$ model--question cohorts:
routing similarity is concentrated early, maximally differentiated at
intermediate budgets, and reconcentrated later, and the timing of this maximum
varies systematically with architecture and reasoning effort. An exact
decomposition separates cohort-wide common-mode mass from residual spectral
dimensionality: common-mode reallocation accounts for about two thirds of the
trajectory, while the residual spectrum contributes about one quarter and
retains substantial variation beyond the common mode. The decomposition further
localizes behavior: among non-unanimous cohorts, increases in common-mode
concentration strongly predict same-answer recoverability, and higher reasoning
effort delays the maximum by $2.59$ octaves (doublings of the token budget) and
consistently expands the high-rank period across all four tested architectures,
locating the effort effect in timing and duration rather than peak amplitude.
Correctness comparisons separate structural monitoring from answer selection,
positioning routing effective rank as a decomposable, label-free diagnostic of
cohort organization---a principled spectral lens on how MoE reasoning cohorts
differentiate and reconcentrate over inference time.
}

\begin{document}
\maketitle

\begin{figure}[t]
  \centering
  \includegraphics[width=\textwidth]{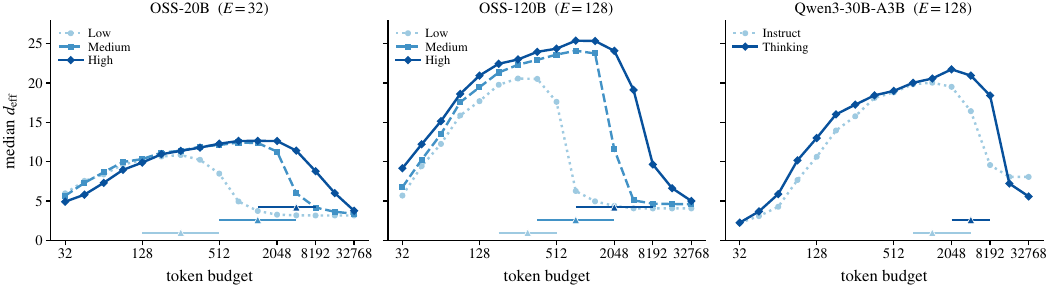}
  \caption{\textbf{Routing effective rank follows a reproducible
  low--high--low trajectory whose peak timing is architecture- and
  effort-dependent.} One panel per architecture (fixed expert count $E$;
  Qwen3-Next-80B, $E{=}512$, and the pooled curve are in
  \Cref{fig:app:famtraj}), one curve per reasoning-effort tier (medians).
  Triangles and whiskers mark the median and IQR of the peak budget, measured
  separately within each cohort (\Cref{sec:rq1}). The maximum moves later with
  effort in all four architectures, whereas peak height is
  architecture-dependent. Panels share one $y$-scale, but absolute $\deff$
  levels depend on the routing representation and cohort size ($\deff\!\le\!N$):
  compare timing and shape within an architecture, not levels across panels.}
  \label{fig:trajectory}
\end{figure}

\section{Introduction}
\label{sec:intro}

Test-time scaling turns inference into a population process: instead of one answer, a
model produces a \emph{cohort} of candidate reasoning trajectories and combines
them by majority vote, a verifier, or a learned
reward~\citep{wang2023selfconsistency,lightman2024lets,snell2024scaling}. As systems
spend more compute at inference, the cohort itself becomes an object of study: it is
what self-consistency aggregates and what group-based training treats as a unit.

Most existing analyses characterize a reasoning cohort through its final
answers---votes, margins, or verifier scores. Before those answers are compared,
the cohort already has an internal organization: some trajectories compute
similarly, others route through different expert patterns. Sparse
Mixture-of-Experts (MoE) models expose this organization through their routing
traces. Routing is a by-product of decoding rather than an added probe: it
arrives with every token and can be read online at any prefix, yet we find it
carries a reproducible signal about how the cohort organizes its reasoning.
Comparing traces across same-question rollouts yields an activation-only
similarity graph $\Wmat$ (\Cref{sec:method}) whose spectrum compactly describes
cohort geometry.

We study the \emph{routing effective rank} $\deff$, the entropy-effective
dimensionality of that routing-similarity spectrum. Unlike ordinary rank, it discounts
numerically tiny directions; unlike graph density, it summarizes the full eigenvalue
distribution. Its interpretation, however, depends on how spectral mass is
allocated: a rise in $\deff$ may reflect a weaker cohort-wide routing mode, a
richer residual spectrum, or both. We therefore study $\deff$ as a decomposable
trajectory rather than an isolated scalar (\Cref{fig:trajectory}), in three
stages: its temporal regularity and architecture-dependent timing (RQ1), the
spectral channels that account for it (RQ2), and how those channels relate to
answer organization and reasoning effort (RQ3).

Across ten MoE configurations and five math/science benchmarks, $\deff$ traces a
robust low--high--low trajectory over token budget, with a prominent interior maximum
in $98.5\%$ of $\Ncohorts$ model--question cohorts (\Cref{fig:trajectory}). The shape
is reproducible, while its timing is systematically architecture- and
effort-dependent: the peak budget spans roughly $5\times$
across families and moves consistently later with reasoning effort. Permuting
expert identities removes the prominent arc, indicating that it reflects
cross-rollout routing organization rather than marginal expert usage, while
difficulty-stratified and still-generating-rollout analyses preserve the
qualitative shape. We therefore treat the trajectory as a reproducible spectral
regularity of the cohort, and ask what produces it.


To explain this regularity, we derive the exact decomposition
$\log\deff=h(m)+(1-m)\log\dperp$, where $m$ is the mass of the leading eigenmode and
$\dperp$ is the effective dimensionality of the residual spectrum. The leading mode
aligns almost perfectly with the cohort-wide agreement direction (squared alignment
$0.98$). Reallocating mass into and out of this mode accounts for roughly two thirds
of the trajectory, while the residual spectrum contributes about one quarter and
retains substantial variation beyond the common mode (grouped-CV
$R^2=0.37$). The arc therefore combines a dominant common-mode
weakening-and-recovery process with a distinct residual reconfiguration.

The decomposition localizes distinct empirical roles: common-mode mass governs
how easily same-answer rollout pairs are recovered from routing alone
(\emph{same-answer recoverability}), reasoning effort delays and widens the
high-rank regime, and residual dimensionality retains complementary structural
variation. Correctness comparisons position the metric as a label-free
diagnostic of cohort organization and temporal routing geometry.

\paragraph{Contributions.}
This work makes three contributions.
{\renewcommand{\labelitemi}{$\bullet$}
\begin{itemize}
\setlength{\itemsep}{1pt}\setlength{\parskip}{0pt}
\item We establish a \textbf{robust routing
differentiation--reconcentration trajectory} across ten MoE configurations, five
math/science benchmarks, and $\Ncohorts$ model--question cohorts, validated by
expert-identity, difficulty-stratified, and still-generating controls (RQ1).
\item We derive an \textbf{exact common-mode/residual decomposition} of routing
effective rank and show that common-mode mass allocation accounts for most of its
temporal motion, while residual dimensionality retains substantial independent
variation (RQ2).
\item We \textbf{behaviorally localize the two channels}:
common-mode concentration predicts same-answer recoverability, and higher
reasoning effort consistently delays and broadens the high-rank regime across four
architectures (RQ3).
\end{itemize}}
Together, these results turn $\deff$ from a visually
suggestive curve into an interpretable, testable measurement framework for MoE
cohort dynamics.

\section{Related Work}
\label{sec:related}

\paragraph{Test-time cohorts, internal signals, and multi-run geometry.}
Sampling and aggregating many reasoning trajectories is standard, through majority
vote~\citep{wei2022chain,wang2023selfconsistency}, outcome or process
verifiers~\citep{cobbe2021gsm8k,uesato2022solving,lightman2024lets}, compute
allocation~\citep{snell2024scaling,brown2024monkeys,muennighoff2025s1}, group-based
training~\citep{shao2024deepseekmath}, adaptive
stopping~\citep{aggarwal2023adaptive,li2024escape}, and confidence or selective
prediction~\citep{geifman2017selective,kadavath2022know,kuhn2023semantic}. A
white-box line instead builds selectors and verifiers from internal computation:
cross-rollout neuron agreement and novelty~\citep{chen2025nad,chen2026nex},
hidden-state probes and temporal
signals~\citep{ni2026reprobe,vilas2025tracing,zhang2026telltale}. Traces have also
been mapped as landscapes, graphs, topological signatures, and representation-space
trajectories~\citep{zhou2025landscape,xiong2025mapping,lee2026reasoningflow,tan2025shape,sun2026trajectories};
closest in structure, SliceGraph links activation slices across sampled chains into
mutual-$k$NN process families~\citep{chen2026slicegraph}. We instead treat the
cohort as one unlabeled measurement object---a whole-rollout graph with
routing-similarity edges, read as a budget-indexed spectrum---using answer-side
signals to localize behavioral coupling and define the intended use.

\begin{figure*}[t]
  \centering
  \includegraphics[width=5.5in]{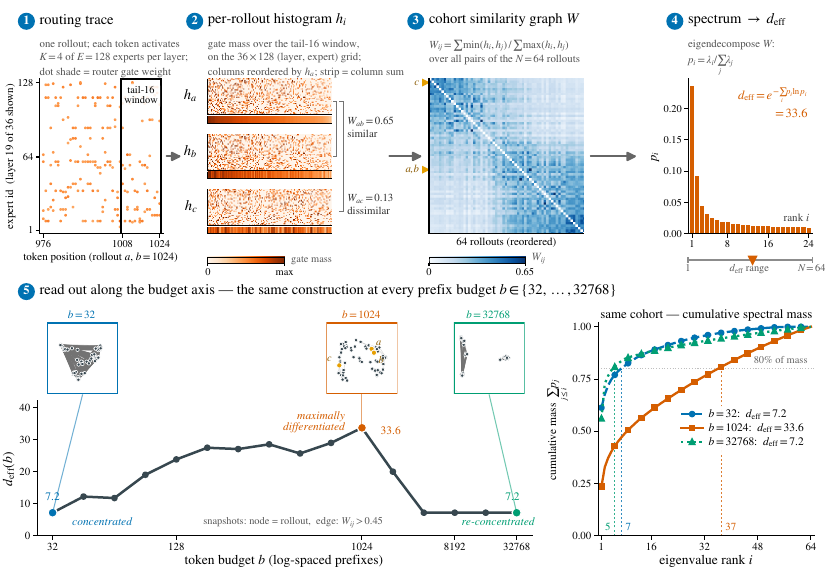}
  \caption{\textbf{From routing traces to $\deff$, on one real cohort
  (OSS-120B-High, AIME-24).} \textbf{(1--2)}~Each rollout's routing trace
  ($K{=}4$ of $E{=}128$ experts per token per layer) is summarized, over a
  trailing $16$-token window at budget $\bbudget$, by a gate-weight histogram
  $h_i$---from router activations alone; answer annotations only locate the
  commit window and label alignment.
  \textbf{(3)}~Weighted-Jaccard similarities (\Cref{eq:wj}) form the cohort
  graph over $N{=}64$ rollouts. \textbf{(4)}~Its normalised spectrum gives
  $\deff$ (\Cref{eq:deff}), from $1$ to $N$. \textbf{(5)}~Read at every budget,
  $\deff$ traces the low--high--low trajectory ($7.2\to33.6\to7.2$; the ends need
  $5$--$7$ eigendirections for $80\%$ of spectral mass, the peak $37$): routing
  similarity is first concentrated, then distributed across more spectral
  directions, and finally reconcentrated---a direct geometric readout of cohort
  organization.}
  \label{fig:defchain}
\end{figure*}

\paragraph{MoE routing as an internal signal.}
Sparse MoE layers~\citep{shazeer2017moe,lepikhin2021gshard,fedus2022switch} underpin
many current models~\citep{jiang2024mixtral,dai2024deepseekmoe}. Prior work studies
specialization, load balancing, and routing
stability~\citep{zoph2022stmoe,zoph2022designing}, prefetching~\citep{eliseev2023fast},
test-time rollout selection~\citep{chen2026rad}, and what routes encode: probing
finds fine-grained expert specialization~\citep{herbst2026expert}, geometric
accounts counter that routing mirrors hidden-state
organization~\citep{wang2026myth}, counterfactual analyses probe route
utility~\citep{yoon2026misrouted,ying2025moemui}. Our question is complementary:
how the cross-rollout routing graph reorganizes over budget.

\paragraph{Effective rank and spectral diversity.}
Effective rank is the exponential of spectral entropy~\citep{roy2007effective},
used statically to diagnose representation collapse~\citep{garrido2023rankme}; the
same functional underlies the Vendi score~\citep{friedman2023vendi} and order-one
Hill diversity~\citep{hill1973diversity}. We study its temporal behavior on routing
similarity graphs, derive an exact common-mode/residual decomposition, and localize
behavioral coupling to those channels; spectral clustering and Nystr\"om completion
appear only as structural controls
~\citep{vonluxburg2007tutorial,williams2001nystrom}.

\section{Method: Routing Effective Rank and Its Spectral Anatomy}
\label{sec:method}


\subsection{Cohorts and the Routing Graph}
\label{sec:setup}
For each question we sample a cohort of $\nrollouts{=}64$ rollouts and evaluate along
a logarithmic \emph{budget axis} of token prefixes
$\bbudget\in\{32,48,\dots,32768\}$ ($16$ points), plus a per-rollout \emph{commit
window} located at each boxed final answer. At budget $\bbudget$, each rollout $i$
has a histogram $h_i(\bbudget)$ accumulating the router's gate weight over each
$(\text{layer},\text{expert})$ slot in a short trailing window ($16$ tokens by
default) (\Cref{fig:defchain}). Edge weights are computed from router activations
only; answer annotations never enter the similarity---they select the boxed-answer
rollout subset, locate the commit window via a fixed answer-marker token anchor,
and label same-answer alignment. The symmetric cohort graph
$\Wmat\in\mathbb{R}^{\nrollouts\times\nrollouts}$ holds their \emph{weighted Jaccard}
similarity,
\begin{equation}
  \Wmat_{ij}(\bbudget) \;=\;
  \frac{\sum_{e}\min\!\big(h_i^{(e)}(\bbudget),\,h_j^{(e)}(\bbudget)\big)}
       {\sum_{e}\max\!\big(h_i^{(e)}(\bbudget),\,h_j^{(e)}(\bbudget)\big)},
  \label{eq:wj}
\end{equation}
indexed over expert slots $e$; $\Wc$ denotes the commit-window graph. Two companion
readouts recur: \emph{density} $\dens$, the off-diagonal mean of $\Wmat$ (plain
routing agreement), and \emph{same-answer AP}, the average precision of ranking
same-final-answer rollout pairs by edge weight. Two derived readouts recur in RQ3:
the answer effective count $\dans=\exp(-\sum_c\pi_c\log\pi_c)$ over the empirical
answer-cluster distribution (the same Hill functional as $\deff$), and the
prevalence-corrected
$\mathrm{AP}_{\mathrm{lift}}=(\mathrm{AP}-\pi_{\mathrm{same}})/(1-\pi_{\mathrm{same}})$,
with $\pi_{\mathrm{same}}$ the same-answer pair prevalence.

\subsection{Routing Effective Rank}
\label{sec:deff}
With $p_i=\lambda_i/\sum_j\lambda_j$ over the eigenvalues of the symmetrized
$\Wmat$, the effective rank is the exponential of the spectral
entropy~\citep{roy2007effective},
\begin{equation}
  \deff(\Wmat) \;=\; \exp\!\Big(-\textstyle\sum_{i} p_i \log p_i\Big),
  \label{eq:deff}
\end{equation}
the entropy-effective dimensionality of the routing-similarity spectrum---the same
functional as the Vendi score on similarity matrices~\citep{friedman2023vendi}. Under
approximately block-structured geometry it approximates an effective number of
balanced routing groups; more generally, it quantifies the spectral dimensionality
of cohort routing similarity rather than a literal count of clusters or solutions.
These matrices are PSD to numerical precision (\Cref{sec:app:psd}); we symmetrize and
clip negative eigenvalues before the entropy. The absolute scale is
representation-specific, so we read $\deff$ by rank. In plain terms, low $\deff$
means that a few shared patterns explain most of the routing graph; high $\deff$
means that more independent directions are needed.

\subsection{An Exact Spectral Decomposition}
\label{sec:decomp}
Effective rank can change because the leading eigenmode gains or loses mass, because
the residual spectrum expands or contracts, or because mass mixes between the two.
The grouping property of Shannon entropy separates these channels exactly. Let
$m=p_1$ be the normalised mass of the leading eigenvalue, let $q_i=p_i/(1-m)$ for
$i\ge2$ renormalise the rest, and let
$\dperp=\exp\big(-\sum_{i\ge2}q_i\log q_i\big)$ be the effective dimensionality of
the residual spectrum. With $h(m)=-m\log m-(1-m)\log(1-m)$,
\begin{equation}
  \log \deff \;=\; h(m) \;+\; (1-m)\,\log \dperp
  \label{eq:decomp}
\end{equation}
holds to machine precision wherever $\dperp$ is defined (numerical accounting in
\Cref{sec:app:anatomy}).
Writing $r=\log\dperp$, any trajectory segment is attributed exactly to its three
channels by the midpoint split
$\Delta\log\deff=\Delta h(m)-\bar r\,\Delta m+(1-\bar m)\,\Delta r$. We call $m$ the
\emph{common-mode mass}---RQ2 shows the leading eigenvector aligns almost perfectly
with the cohort-wide agreement direction---and use $r=\log\dperp$ for the residual
dimensionality.

\subsection{Statistical Protocol}
\label{sec:protocol}
The unit of analysis is one (model, question) cohort; budget steps within a cohort
are never treated as independent samples. Headline intervals are two-way clustered
bootstraps (question $\times$ configuration, $B{=}2000$ for the confirmatory
families), and
predictive comparisons use $5$-fold cross-validation grouped by question, so the
same question never straddles train and test. RQ3's three confirmatory endpoints,
decision rule, and verdict sentence were fixed before the analyses ran and are
Bonferroni-corrected ($\alpha=0.05/3$); its behavioral-localization layer carries
secondary trajectory endpoints in the same confirmatory statistics, and null
results are stated as effect-size bounds rather than proof of exact zero.
Per-estimate bootstrap schemes, measurement conventions (peak definitions,
prominence gates, grid endpoints), the full pre-registered plan, and the decision
record are in \Cref{sec:app:protocol} and \Cref{sec:app:audit}.

\section{Experiments}
\label{sec:experiments}
RQ1 establishes the temporal phenomenon, RQ2 reveals its spectral anatomy, and
RQ3 localizes its behavioral associations.
\subsection{Experimental Setup}
\label{sec:exp:setup}

We use ten MoE configurations---OSS-\{20B,120B\} at High/Med/Low reasoning
effort~\citep{openai2025gptoss} ($E{=}32/128$, $K{=}4$; one base checkpoint per
scale, effort set in the system prompt), Qwen3-30B-A3B-\{Instruct,Thinking\} ($E{=}128$,
$K{=}8$), and Qwen3-Next-80B-A3B-\{Instruct,Thinking\}~\citep{yang2025qwen3}
($E{=}512$, $K{=}10$)---on five math/science benchmarks
(AIME'24/'25~\citep{maa_aime}, BRUMO'25~\citep{brumo2025}, HMMT'25~\citep{hmmt},
GPQA~\citep{rein2024gpqa}). This
yields $\Ncohorts$ model--question cohorts ($\nrollouts{=}64$ rollouts each) over $50$
model$\times$dataset shards; per-analysis retention filters are accounted for in
\Cref{tab:app:accounting}. Configurations group into three families---\textsf{OSS},
\textsf{Qwen-Instruct}, \textsf{Qwen-Thinking}---but family co-varies with
routing architecture (expert count, top-$K$, depth, training recipe), so
cross-family comparisons are reported descriptively, as rank-based trajectory
contrasts. A sixth benchmark, LiveCodeBench-v5~\citep{jain2025livecodebench}, was held
out from every analysis decision and used once, with all dials frozen
(\Cref{sec:app:transfer}).

\subsection{RQ1: How Does Routing Effective Rank Evolve During Reasoning?}
\label{sec:rq1}

We first characterize the trajectory at the level of routing geometry. Pooled
over five math and science benchmarks, $\deff$ rises from $5.21$ at the earliest
budget to $18.68$ at $\bbudget{=}512$ and returns to $5.84$ at the largest budget
($5.76$ at commit; \Cref{fig:app:famtraj}, inset). In spectral terms, routing
similarity is concentrated early, spread over more directions at intermediate
budgets, and concentrated again later (\Cref{fig:app:spectra}).

\paragraph{The pattern appears within individual cohorts.}
A prominent interior maximum occurs in $98.5\%$ of the $\Ncohorts$ model--question
cohorts ($95\%$ CI $[96.7,99.8]$, two-way clustered by question and configuration),
with a median relative prominence of $0.68$ of the peak height. The rate is $97.9$--$99.3\%$ on every
benchmark and ${\ge}92.9\%$ on every configuration, and the same qualitative form
appears in every difficulty stratum ($98.3$--$98.7\%$), though peak height and
location move with difficulty (\Cref{sec:app:protocol}). The arc is therefore
present at the individual-cohort level, beyond the pooled average.

\paragraph{Peak timing varies across models, datasets, and peak definitions.}
Where the maximum falls depends on the model and on how the maximum is measured.
Locating the maximum after averaging a family's curves gives earlier peaks than
locating each cohort's maximum and taking the median---the two definitions differ
by up to $1.5$--$2$ octaves (doublings of the token budget), because averaging
first is pulled early by the spread of individual peak positions. Under the
per-cohort definition, used throughout and declared per figure (the ``caliper'' of
\Cref{sec:app:protocol}), the median peak budgets are $768$ tokens for
\textsf{OSS}, $1024$ for \textsf{Qwen-Instruct}, and $4096$ for
\textsf{Qwen-Thinking}. Under every definition, the pooled peak at
$\bbudget{=}512$ summarizes a mixture of family-specific peak distributions;
per-cohort medians provide the appropriate within-family timing statistic.
Across datasets, by contrast, the
\emph{ordering} of peak timing is nearly fixed---almost identical in all ten
configurations (Kendall $W{=}0.946$), with GPQA peaking $1.5$--$2$ octaves
before the math benchmarks (\Cref{fig:app:datasetclock}).

\paragraph{The pattern depends on shared expert identities across rollouts.}
Permuting expert identities within each rollout and layer preserves per-rollout
activation counts, per-layer load, and sparsity exactly, and destroys only the
alignment of expert identities \emph{across} rollouts. This control reduces the
prominent-peak rate from $98.5\%$ to $0.0\%$ ($10/10$ configurations), cuts median
prominence by a factor of $178$, and flattens the spectrum to $\deff\!\approx\!n$
(median ratio $0.94$ to cohort size); a formally defined argmax survives on $73\%$
of the near-flat curves, so the control removes the peak's magnitude, not the
existence of a maximum on a noisy curve. Restricting the graph to still-generating
rollouts (the risk-set control of \Cref{sec:app:controls}) preserves an interior
maximum in $99.8\%$ of cells, with a peak location correlated at $\rho=0.63$ with
the full graph: completed traces refine the exact late-stage timing while leaving
the trajectory intact. Together, these controls establish the arc as a property
of aligned cross-rollout routing organization rather than of marginal activation
volume.

\paragraph{Temporal context disambiguates equal-rank states.}
The same effective-rank value can occur on both the rising and the falling branch,
and the mean spectra at the two endpoints are nearly identical ($p_1$ differs by
$0.001$; \Cref{fig:app:spectra}). This branch symmetry makes $\deff$ most
informative as a trajectory-valued diagnostic: budget context and local trend
distinguish routing differentiation from reconcentration, so effective rank
captures both cohort geometry and its temporal organization when read along the
budget axis.

\takeaway{1}{Routing effective rank reveals a robust
differentiation--reconcentration trajectory whose maximum shifts systematically
with architecture and reasoning effort, establishing temporal organization as an
architecture-aware property of MoE reasoning cohorts.}

\subsection{RQ2: What Drives the Effective-Rank Trajectory?}
\label{sec:rq2}

\begin{figure*}[t]
  \centering
  \includegraphics[width=\textwidth]{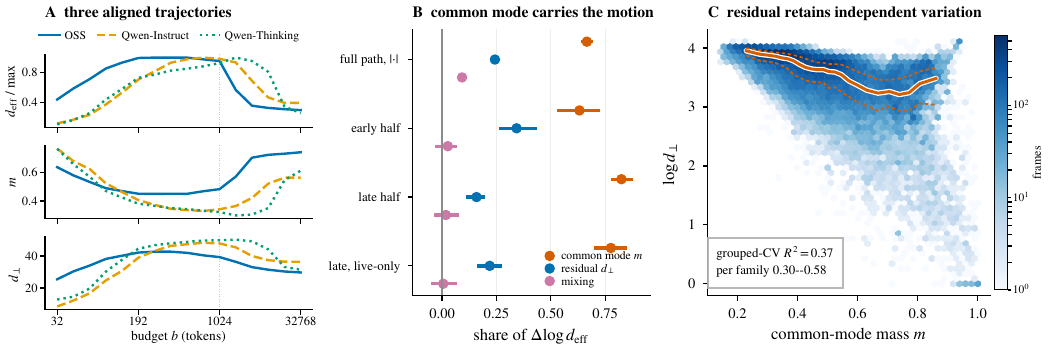}
  \caption{\textbf{Spectral anatomy of routing effective rank.} The identity
  $\log\deff = h(m) + (1-m)\log\dperp$ is exact, with $m=p_1$ the common-mode mass
  and $\dperp$ the residual dimensionality (evaluated term by term on a real
  eigenspectrum in \Cref{fig:app:identity}). \textbf{(A)} Aligned per-family
  medians: as $\deff$ rises, spectral mass drains out of the common mode; as it
  falls, mass returns. The residual spectrum contributes complementary motion on
  a smaller scale.
  \textbf{(B)} Over the full budget path the common-mode term carries $0.67$ of
  the absolute motion of $\log\deff$, the residual term $0.24$, and mixing $0.09$
  (two-way clustered $95\%$ CIs); the remaining rows repeat the accounting on the
  early and late halves of the budget grid (details and controls in
  \Cref{sec:app:anatomy}). \textbf{(C)} Residual dimensionality retains
  substantial variation beyond the common mode: a grouped cross-validated fit of
  $\log\dperp$ on $m$ explains $37\%$ of out-of-sample variance, leaving $63\%$
  as a distinct spectral axis for RQ3.}
  \label{fig:anatomy}
\end{figure*}

RQ1 establishes a reproducible low--high--low trajectory. \Cref{eq:decomp}
resolves its spectral source by separating leading-mode mass, residual
dimensionality, and their interaction, and attributes every trajectory segment
exactly to these channels. \Cref{fig:anatomy} summarizes the result.

\paragraph{The leading mode is a common mode.}
Its eigenvector is almost exactly the graph-wide agreement direction:
$|\langle v_1,u\rangle|^2$ has median $0.9838$ $[0.9800,0.9872]$, above $0.8$ in
$98.3\%$ of frames and above $0.98$ in every model family (intervals two-way
clustered by question and configuration). Alignment is lowest precisely at the effective-rank
maximum ($0.970$ vs.\ $0.991$ at the ends), and $m$ tracks plain graph density within
a cohort ($\rho=+0.987$): the leading mode \emph{is} routing agreement, in spectral
form.

\paragraph{Common-mode mass allocation carries most of the motion.}
Over the $16$-budget grid the absolute-value-normalised shares are $0.666$
$[0.645,0.684]$ for the common-mode term, $0.243$ $[0.230,0.260]$ for residual
dimensionality and $0.092$ for mixing, positive in $10/10$ configurations and
unchanged on a linear $\deff$ scale. Splitting the grid at its log-midpoint, the common-mode term
takes $0.63$ of the early half and $0.82$ of the late half, and exceeds the residual
term in both (Bonferroni-corrected over the three pre-registered hypotheses;
\Cref{sec:app:anatomy}). These shares quantify each channel's contribution to
$\Delta\log\deff$ under the exact identity: normalised by each component's
available range, the two channels move comparably fast in the early half, so the
common mode's larger share reflects its greater spectral leverage---the
$5.9\times$ difference in the identity's multipliers.

\paragraph{Channel attribution is stable across the budget axis.}
Permuting each problem's peak position across problems reproduces most of the
attribution, and every fixed cut point gives a similar split: the accounting is a
property of the budget axis rather than of the estimated peak. The residual share
is larger early than late ($+0.158$ $[+0.080,+0.232]$; matched-cell contrast), and
restricting the late
window to still-generating rollouts attenuates the contrast by $45\%$ while
leaving it positive. A matched permuted-peak placebo attributes most of the
apparent near-peak residual enrichment to budget position, which sharpens the
supported claim: attribution is stable along the budget axis rather than driven
by a peak-localized burst (full numbers in \Cref{sec:app:anatomy}).

\paragraph{A sparsity-preserving expert-identity control.}
Permuting expert identities per rollout and layer---preserving sparsity, load and
every marginal activation statistic---drives the residual spectrum toward its
near-isotropic ceiling ($\dperp=62.9$ against a ceiling of $63$) and leaves a
prominent interior peak in $3.4\%$ of cells against $97.5\%$ for real cohorts (this
experiment's own pool and gate; \Cref{sec:app:protocol} explains why peak rates are
never compared across gates). The decomposition therefore measures \emph{which}
experts fire together rather than how much routing mass moves. Cross-problem
pseudo-cohorts, by contrast, retain the coarse arc in $97.0\%$ of $264$ cells,
revealing a generation-level temporal backbone; together, the two controls
separate this shared backbone from the aligned routing organization measured
within cohorts.

\paragraph{Residual dimensionality forms a complementary axis.}
Although $m$ and $r=\log\dperp$ are mathematically distinct, their empirical
relationship is also far from one-dimensional: a grouped cross-validated fit of
$\log\dperp$ on $m$ explains $37\%$ of out-of-sample variance, leaving $63\%$
beyond common-mode mass ($25\%$ explained on the descending branch of the
\textsf{OSS} family). Residual dimensionality therefore provides a distinct
coordinate of the routing graph, motivating the behavioral localization in RQ3.

\takeaway{2}{The trajectory is driven primarily by spectral mass leaving and
later returning to a cohort-wide common mode. Residual dimensionality accounts
for a complementary share of the motion and retains $63\%$ of its out-of-sample
variation beyond common-mode mass, providing a second, distinct coordinate of
cohort routing geometry.}

\subsection{RQ3: How Do Spectral Channels Relate to Answer Organization and Reasoning Effort?}
\label{sec:rq3}

\begin{figure*}[t]
  \centering
  \includegraphics[width=0.68\textwidth]{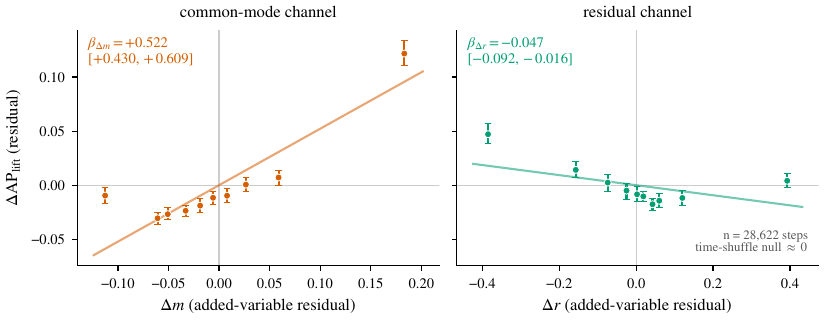}\\[-2pt]
  \includegraphics[width=0.72\textwidth]{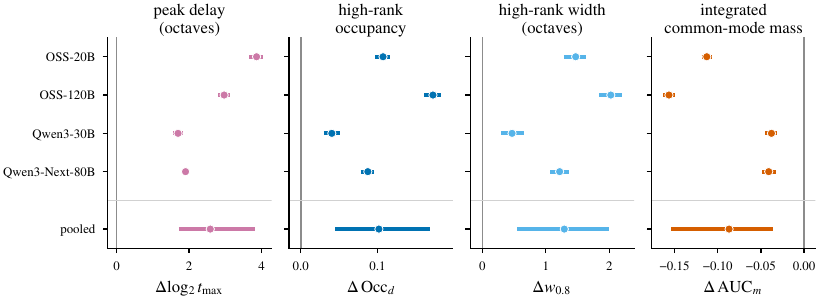}
  \caption{\textbf{Common-mode concentration predicts same-answer
  recoverability; reasoning effort expands the high-rank regime.}
  \textbf{Top:} across $\Nsteps$ adjacent-budget steps (clustered
  on $318$ questions $\times$ $10$ configurations), increases in common-mode mass
  make same-answer rollout pairs easier to recover from routing (the
  prevalence-corrected AP lift plotted here): $\beta_{\Delta m}=+0.522$
  $[+0.430,+0.609]$, while the corresponding residual coefficient is
  $\beta_{\Delta r}=-0.047$ $[-0.092,-0.016]$; a within-question time shuffle
  collapses both toward zero. \textbf{Bottom:} higher effort delays the maximum,
  increases high-rank occupancy, widens the $\deff\!\ge0.8d_{\max}$ regime, and
  lowers integrated common-mode mass, with the same sign in all four
  architectures.}
  \label{fig:behavior}
\end{figure*}

RQ2 yields two mathematically distinct spectral channels; this section, the
paper's behavioral-localization layer (\Cref{sec:protocol}), localizes their
empirical roles. Common-mode concentration is strongly associated with
same-answer recoverability; reasoning effort consistently controls the timing and
duration of the high-rank regime; and residual dimensionality provides a
complementary structural axis. Together, these confirmatory analyses identify
where the routing spectrum is most behaviorally informative.

\paragraph{Common-mode mass tracks same-answer recoverability.}
We measure \emph{same-answer recoverability}---how well routing similarity alone
recovers which rollouts share the same final answer---by the prevalence-corrected
AP lift of \Cref{sec:setup}, among non-unanimous cohorts. In a dynamic model
$\Delta\mathrm{AP}_{\mathrm{lift}}\!\sim\!\Delta m+\Delta r+$ budget-step fixed
effects, increases in common-mode mass strongly predict recoverability gains
(\Cref{fig:behavior}, top). The residual partial is opposite-signed and about a
quarter the size per standardized step, while a within-question temporal shuffle
collapses both coefficients toward zero. The magnitude contrast identifies the
shared spectral mode as the channel through which routing similarity becomes
answer-aligned; its near-equivalence to graph density ($\rho{=}0.987$) gives
that familiar signal a precise spectral interpretation.

\paragraph{Reasoning effort expands the high-rank regime.}
Pairing effort tiers within architecture yields four directionally consistent
effects (\Cref{fig:behavior}, bottom): higher effort delays the maximum by
$+2.59$ octaves $[+1.72,+3.83]$, raises peak-normalized occupancy---the time
spent near high effective rank---by $+0.102$ $[+0.045,+0.168]$, widens the
$\deff\!\ge0.8d_{\max}$ interval by $+1.29$ octaves $[+0.55,+1.99]$, and reduces
integrated common-mode mass by $-0.087$ $[-0.154,-0.036]$, while residual-spectrum
area remains comparatively stable. The registered peak-height contrast varies
across architectures and its pooled interval spans zero ($+0.104$
$[-0.079,+0.232]$), whereas the timing and duration readouts agree directionally
in all four
(\Cref{tab:app:effort}). This contrast localizes the robust effort effect to
temporal extent
rather than peak amplitude: higher effort sustains a less
common-mode-concentrated, high-rank routing regime for longer, and peak-aligned
curves retain architecture-specific shapes---broadening beyond a rigid temporal
translation.

\paragraph{Behavioral localization and complementary use.}
Under the pre-registered decision tree (\Cref{sec:app:audit}), the tested
answer-space associations concentrate primarily in the common-mode channel, while
residual dimensionality remains the distinct structural coordinate established in
RQ2. Correctness comparisons separate two roles: answer-side statistics remain
stronger for answer selection
(\Cref{tab:app:correctness}), while routing reaches up to $0.68$ within a problem
(\Cref{sec:app:boundary}) and exposes label-free cohort geometry. We therefore position $\deff$ as a label-free monitor
of cohort geometry and temporal organization, with common-mode mass as its most
directly answer-aligned channel.

\takeaway{3}{Common-mode concentration is the behaviorally aligned spectral
channel: it strongly predicts same-answer recoverability, while higher reasoning
effort consistently delays and broadens the high-rank regime. Residual
dimensionality contributes complementary structural variation, and the
decomposition localizes the operational content of $\deff$ instead of treating it
as an undifferentiated scalar.}

\section{Robustness, Scope, and Limitations}
\label{sec:limitations}

Our claims concern within-architecture trajectory shape and timing, because the
absolute value of $\deff$ depends on the routing representation, expert count, and
cohort size. The core low--high--low trajectory is robust across complementary
controls: restricting to still-generating rollouts preserves the interior
maximum; masking the boxed answer leaves the trajectory statistically unchanged
(\Cref{sec:app:controls}); a frozen transfer to $1{,}582$ LiveCodeBench-v5
problems yields a $97.7\%$ prominent-peak rate (\Cref{sec:app:transfer}); and a
binary expert-set Jaccard graph preserves the geometric results while slightly
improving same-answer AP (\Cref{sec:app:binary})---so the diagnostic can be
computed from sparse expert-activation indicators alone, without router gate
magnitudes.

These controls also identify where calibration matters. Risk-set restriction
refines exact late-stage timing ($\rho=0.63$ with the full graph), and peak-frame
estimates require larger subcohorts than terminal ones
(\Cref{sec:app:subcohort}). Lexical and formatting cues may remain partially
entangled with same-answer recoverability, and sensitivity to sampling
temperature remains to be characterized.

The confirmatory behavioral analysis rests on $318$ unique questions under
two-way clustering, so the reported intervals define the resolution of the
present effect-size claims, and cross-family results are read as rank-based
trajectory comparisons rather than capability claims.

\section{Conclusion}
\label{sec:conclusion}

We introduced routing effective rank as a label-free spectral view of how
same-question MoE reasoning cohorts reorganize over inference time. Across ten
configurations and five math/science benchmarks it traces a reproducible
differentiation--reconcentration trajectory whose timing shifts with
architecture and reasoning effort.

An exact decomposition shows that common-mode mass allocation drives most of
this motion, while residual dimensionality retains substantial independent
structure. It also localizes behavior: common-mode concentration predicts
same-answer recoverability among contested cohorts, and higher reasoning effort
delays and broadens the high-rank regime across all four tested architectures.

The measurement is also practical: the trajectory and its readouts survive a
frozen transfer to held-out code generation, and the diagnostic can be computed
from sparse expert-activation indicators alone, so the framework applies during
serving, before any answer is graded.

These results establish $\deff$ as an architecture-aware,
label-free diagnostic of cohort routing geometry and temporal
organization---turning a visually suggestive curve into a decomposable
measurement framework for tracking when MoE reasoning cohorts differentiate,
sustain distributed routing structure, and reconcentrate as inference compute
unfolds.

\small
\bibliographystyle{plainnat}
\bibliography{references}
\normalsize

%
%
\clearpage
\appendix
\label{aaai:appendixstart}
\raggedbottom
\setcounter{dbltopnumber}{2}
\setcounter{figure}{0}\renewcommand{\thefigure}{A\arabic{figure}}
\setcounter{table}{0}\renewcommand{\thetable}{A\arabic{table}}
\setcounter{equation}{0}\renewcommand{\theequation}{A\arabic{equation}}

\section*{Technical Appendix}
\suppressfloats[t]
\noindent
This appendix collects the protocol declarations, tables, controls, sensitivity
analyses, and supplementary results referenced from the paper. Its sections are lettered A, B, \dots{} and its
figures, tables, and equations are numbered A1, A2, \dots{}, so a pointer from the
paper reading ``\Cref{sec:app:psd}'' or ``\Cref{tab:app:capability}'' identifies them
unambiguously. Citations here refer to the reference list of the paper.


\section{Statistical Protocol, Calipers, and Gates}
\label{sec:app:protocol}

This section specifies the measurement conventions behind every headline number,
making the analysis fully reproducible.

\paragraph{Unit and clustering.}
The unit of analysis is one (model, question) cohort ($\Ncohorts$ in total), where
``model'' ranges over the ten served configurations; budget
steps within a cohort are never treated as independent samples. Headline intervals
are two-way clustered bootstraps over questions and configurations
($B{=}2000$)~\citep{cameron2011robust}, because this design has at most one cohort
per (question, configuration) pair. Two-way clustering materially affects the
uncertainty estimate: for the
prominent-peak rate ($0.9852$), the iid interval is $[0.9807,0.9894]$ (width
$0.0087$), one-way by question $[0.9803,0.9897]$ ($0.0094$), and two-way
$[0.9671,0.9984]$ ($0.0313$)---$3.6\times$ wider than iid, with the configuration
dimension doing the work. We always report the two-way interval, and each headline
carries the per-configuration sign count ($k/10$: how many of the ten
configurations share the pooled sign). Two estimator families
deviate from this default and are labelled where they appear: the pooled effort
contrasts of \Cref{tab:app:effort} cluster on question $\times$ architecture (four
architecture clusters; per-architecture rows cluster on question only), and the
dynamic recoverability supplement and the correctness-boundary AUCs use $B{=}500$, the
latter uncorrected by design. The configuration dimension's bootstrap clusters are
the ten served configurations; the \textsf{OSS} tiers within a scale share base
weights and differ only in the prompt-set effort level (\Cref{sec:exp:setup}), so
this dimension has six distinct checkpoints behind its ten clusters.

\paragraph{Peak calipers.}
``Where the trajectory peaks'' depends on how the maximum is measured, and the
calipers differ by up to two octaves (\Cref{tab:app:calipers}): the argmax of a
family's \emph{mean} curve, of its \emph{median} curve, and the median of
\emph{per-cohort} argmax positions are all defensible and all different, because
aggregating curves before locating the maximum is pulled early by the spread of
individual peak positions. The paper reports the per-cohort median throughout, and
every figure states its caliper.

\begin{table}[t]
  \centering
  \caption{Three peak-budget calipers: argmax of the family \emph{mean} curve, of
  the family \emph{median} curve, and the median of \emph{per-cohort} argmax
  positions [IQR]. The paper's primary caliper is the per-cohort median (rightmost);
  the family-mean caliper is shown because aggregated curves are what a reader sees
  in a plot.}
  \label{tab:app:calipers}
  \footnotesize
  \setlength{\tabcolsep}{2pt}
  \begin{tabular}{lcccc}
    \toprule
    Family & $n$ & mean & median & per-cohort [IQR] \\
    \midrule
    \textsf{OSS}           & $1863$ & $256$  & $192$  & $768\,[256,2048]$ \\
    \textsf{Qwen-Instruct} & $636$  & $768$  & $768$  & $1024\,[768,4096]$ \\
    \textsf{Qwen-Thinking} & $606$  & $2048$ & $2048$ & $4096\,[2560,8192]$ \\
    \midrule
    pooled                 & $3105$ & $512$  & ---    & $1024$ \\
    \bottomrule
  \end{tabular}
\end{table}

\paragraph{Prominence gates.}
A cohort counts as having a \emph{prominent} interior maximum when its peak clears a
fixed relative-prominence threshold ($1.2\times$), frozen before these analyses and
not itself pre-registered; an \emph{interior} maximum requires only that the argmax
is not an endpoint ($99.3\%$ of cohorts). With
$d_{\mathrm{ends}}=\max(d_{b_1},d_{b_{16}})$ the two grid endpoints' larger value,
we report \emph{absolute} prominence $d_{\max}-d_{\mathrm{ends}}$ (in $\deff$
units) and \emph{relative} prominence $(d_{\max}-d_{\mathrm{ends}})/d_{\max}$; the
$1.2\times$ gate requires an interior argmax with
$d_{\max}\ge1.2\,d_{\mathrm{ends}}$. The spectral-anatomy experiment of
\Cref{sec:app:anatomy} uses its own, more permissive gate on its own $6$-shard pool.
Because the same null data can read $0.0\%$ under one gate and a few percent under
another, \textbf{peak rates are never compared across experiments with different
gates or grids}; each is reported against its own real-data arm.

\paragraph{Grid left edge.}
Peak rates are also sensitive to where the budget grid starts, because the
prominence gate compares against the larger of the two grid endpoints, so the left
edge is part of the measurement convention. On the $16$-budget grid
($\bbudget\ge32$) the pooled prominent rate is $98.5\%$; on a $14$-budget grid
($\bbudget\ge64$) it is $95.3\%$, with the loss concentrated in configurations already
differentiated by $\bbudget{=}64$ (GPQA $98.4\%\!\to\!93.5\%$; OSS-20B-Low
$95.3\%\!\to\!78.9\%$; OSS-120B-Low $97.2\%\!\to\!88.7\%$). Peak-rate numbers are
comparable only at matched grids.

\paragraph{Difficulty strata.}
With the fixed majority-share cutpoints $[0.60,0.875]$, the prominent-peak rate is
$98.3\%$ $[94.0,100]$ (hard, $n{=}638$), $98.7\%$ $[96.1,100]$ (mid, $n{=}622$), and
$98.6\%$ $[96.4,99.9]$ (easy, $n{=}1{,}844$); the three strata total $3{,}104$
because one cohort produced no gradeable answer and therefore has no defined
majority share. The peak's height and location both
move (median peak $\deff$ $21.63/23.10/24.70$; median peak budget $2048/2048/1024$).
The qualitative shape persists across difficulty strata, while peak height and
timing vary systematically with difficulty.

\section{Per-Configuration Levels and Trajectory-Aligned Readouts}
\label{sec:app:capability}

\begin{figure*}[t]
  \centering
  \includegraphics[width=0.75\textwidth]{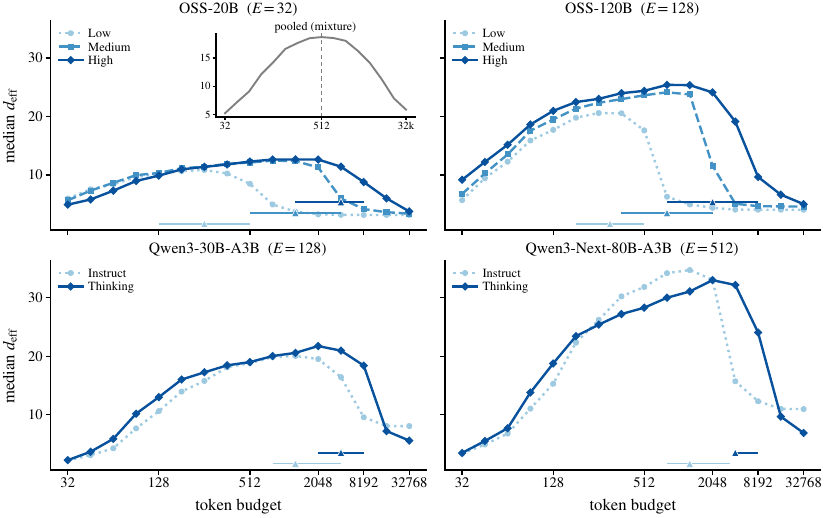}
  \caption{\textbf{Full four-architecture version of \Cref{fig:trajectory},
  including Qwen3-Next-80B ($E{=}512$).} Same data, calipers, and conventions:
  triangles give the per-cohort argmax median with IQR whiskers; the inset is
  the pooled mixture curve. Qwen3-Next-80B repeats the low--high--low shape
  with a later clock and is the one architecture whose peak height inverts
  with effort (per-cohort median peak $\deff$: Instruct $37.2$ vs.\ Thinking
  $36.7$), while its fixed-budget levels remain effort-monotone
  (\Cref{tab:app:capability}).}
  \label{fig:app:famtraj}
\end{figure*}

\begin{figure*}[t]
  \centering
  \includegraphics[width=0.75\textwidth]{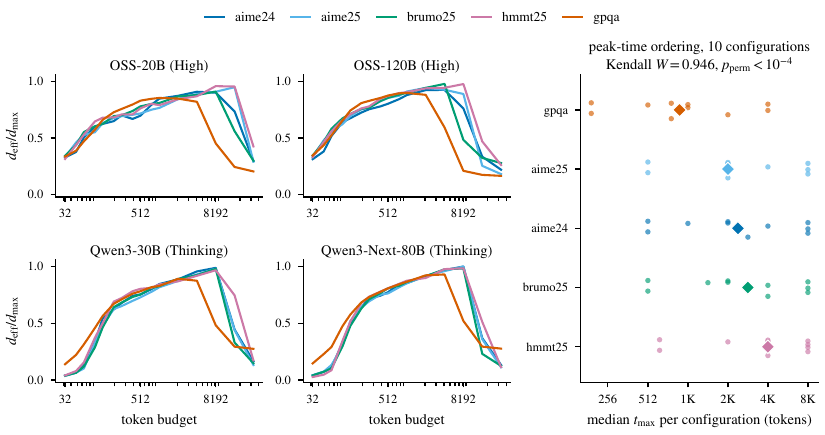}
  \caption{\textbf{The dataset sets the ordering of the effective-rank clock,
  nearly deterministically across configurations.} Left: median
  peak-normalized $\deff$ trajectories per dataset for one representative
  configuration per architecture (High/Thinking tier). Right: each
  configuration's median peak budget per dataset (dots; all ten
  configurations) with cross-configuration medians (diamonds). The dataset
  ordering of peak timing is almost identical in every configuration (Kendall
  $W{=}0.946$, within-configuration permutation $p{<}10^{-4}$): GPQA peaks
  $1.5$--$2$ octaves earlier than the math benchmarks (pooled median
  $\log_2 t_{\max}$ $9.79$ vs.\ $12.00$ for HMMT'25), and HMMT'25 is latest
  with the widest high-rank period ($w_{0.8}$, octaves at
  $\deff{\ge}0.8\,d_{\max}$: median $3.64$ vs.\ $2.38$)---consistent with the
  peak tracking each dataset's length scale rather than a universal token
  count.}
  \label{fig:app:datasetclock}
\end{figure*}

\Cref{tab:app:capability} gives per-configuration $\deff$ levels at a fixed mid-range
budget. Effective rank rises monotonically with reasoning effort inside every
architecture (fixed expert count $E$), isolating effort from model width; absolute
levels are representation- and cohort-size-dependent ($\deff\le N$) and are not
compared across architectures.
\Cref{fig:app:famtraj} completes the trajectory figure of the paper with the
fourth architecture, Qwen3-Next-80B. \Cref{fig:app:datasetclock} adds the
dataset axis: the ordering of peak timing across datasets is nearly invariant
across configurations.

\begin{table}[t]
  \centering
  \caption{Mean/median $\deff$ at $\bbudget{=}4096$ per configuration (weighted
  Jaccard). Monotone in reasoning effort within each architecture.}
  \label{tab:app:capability}
  \small
  \setlength{\tabcolsep}{4pt}
  \begin{tabular}{lccc}
    \toprule
    Configuration & $E$ & mean $\deff$ & median $\deff$ ($n$) \\
    \midrule
    OSS-20B-Low     & $32$  & $3.62$  & $3.19$ ($318$) \\
    OSS-20B-Med     & $32$  & $8.27$  & $5.99$ ($318$) \\
    OSS-20B-High    & $32$  & $10.57$ & $11.41$ ($282$) \\
    OSS-120B-Low    & $128$ & $4.25$  & $4.08$ ($318$) \\
    OSS-120B-Med    & $128$ & $10.62$ & $5.09$ ($318$) \\
    OSS-120B-High   & $128$ & $17.85$ & $19.13$ ($309$) \\
    Qwen3-30B-Instruct    & $128$ & $15.57$ & $16.43$ ($318$) \\
    Qwen3-30B-Thinking    & $128$ & $19.64$ & $20.96$ ($310$) \\
    Qwen3-Next-80B-Instruct & $512$ & $22.04$ & $15.72$ ($318$) \\
    Qwen3-Next-80B-Thinking & $512$ & $29.72$ & $32.18$ ($296$) \\
    \bottomrule
  \end{tabular}
\end{table}

\paragraph{Family means depend on the cohort-size cutoff.} The family fingerprint at
$\bbudget{=}4096$ uses the \emph{matched} cohorts (boxed $\ge17$): $9.2/18.8/25.0$
for \textsf{OSS}/\textsf{Instruct}/\textsf{Thinking}. Pooling the full cache (boxed
$\ge10$) gives $9.1/18.8/24.6$, and a binary-Jaccard graph gives $7.5/14.9/19.6$;
the ordering is invariant, only the absolute scale shifts. The behavioral
counterpart is median commit length ($1537/3133/9690$ tokens).

\begin{table}[t]
  \centering
  \caption{Family mean $\deff$ at a fixed budget, at each family's own trajectory
  peak, and at commit (family-mean caliper; \Cref{sec:app:protocol}). The
  \textsf{OSS}-vs-reasoning separation holds at all three readouts, and the
  \textsf{OSS} effort tiers stay ordered at commit ($3.55/3.65/3.86$ at 20B,
  $4.19/5.26/5.48$ at 120B). The \textsf{Instruct}-vs-\textsf{Thinking} ordering,
  by contrast, is specific to the fixed mid-budget reading: it narrows at the
  peak-aligned readout (\textsf{Thinking} peaks later) and \emph{reverses} at
  commit ($9.2$ vs.\ $6.6$), so fixed-budget readings conflate level with timing.}
  \label{tab:app:phasenorm}
  \small
  \setlength{\tabcolsep}{4pt}
  \begin{tabular}{lccc}
    \toprule
    Family & at $\bbudget{=}4096$ & at own peak & at commit \\
    \midrule
    \textsf{OSS}           & $9.1$  & $15.7$ & $4.3$ \\
    \textsf{Qwen-Instruct} & $18.8$ & $25.1$ & $9.2$ \\
    \textsf{Qwen-Thinking} & $24.6$ & $26.1$ & $6.6$ \\
    \bottomrule
  \end{tabular}
\end{table}

\section{Difficulty Correlations}
\label{sec:app:difficulty}

\Cref{tab:app:difficulty} gives the descriptive difficulty associations referenced
from \Cref{sec:rq3}. We correlate each problem's commit-graph $\deff$ with
\emph{consensus} difficulty ($1-$majority share) and \emph{correctness} difficulty
($1-$accuracy), aggregated two ways: pooling all problems, and averaging over
model$\times$dataset shards (which controls for the cross-model capability axis;
$2000\times$ bootstrap CIs). The pooled-all value is diluted by the capability axis;
the per-shard mean is the interpretable figure.

\begin{table}[t]
  \centering
  \caption{$\mathrm{Spearman}$(commit-graph $\deff$, difficulty). The
  consensus--correctness gap is clear for \textsf{OSS}, ${\approx}0$ for
  \textsf{Instruct}, and reverses for \textsf{Thinking}. Pooled-all is a single
  Spearman over the $3{,}104$ cohorts with a defined majority share (one cohort
  produced no gradeable answer); per-shard mean averages the within-shard Spearman
  over the $50$ model$\times$dataset shards ($2000\times$ bootstrap CIs).}
  \label{tab:app:difficulty}
  \footnotesize
  \setlength{\tabcolsep}{2pt}
  \begin{tabular}{lcc}
    \toprule
    Aggregation & consensus & correctness \\
    \midrule
    pooled-all                         & $+0.15$ & $+0.14$ \\
    \textbf{per-shard mean}            & $\mathbf{+0.46}\,[+.40,+.51]$ & $\mathbf{+0.40}\,[+.34,+.47]$ \\
    \midrule
    per-family: \textsf{OSS}      & $+0.23$ & $+0.12$ \\
    per-family: \textsf{Instruct} & $+0.47$ & $+0.44$ \\
    per-family: \textsf{Thinking} & $+0.27$ & $+0.31$ \\
    \bottomrule
  \end{tabular}
\end{table}

\paragraph{The association is terminal, and window-aligned.}
At $\bbudget\le1024$ every two-way clustered interval covers zero, so we make no
early-triage claim. For the long-reasoning family the readout must be taken at the
commit window: at $\bbudget{=}4096$ it flips negative ($\approx-0.34$) because the
median \textsf{Thinking} commit length is $9{,}690$ tokens, so $\bbudget{=}4096$
still lies on the rising branch there. \textsf{OSS} and \textsf{Instruct}, whose
commits arrive earlier, are already positive at $\bbudget{=}4096$.

\paragraph{Difficulty coupling is likewise concentrated in the common-mode channel.}
Under the pre-registered ladder protocol of \Cref{sec:app:audit}, difficulty as a
regression target behaves like the primaries: adding residual dimensionality after
common-mode mass yields $\Delta R^2_{r\mid m}=+0.022$ $[-0.001,+0.050]$, with the
interval spanning zero, and the split $(m,r)$ state performs comparably to the
scalar. We therefore interpret the descriptive Spearman rows above as difficulty
associations rather than evidence for an additional residual-specific mechanism.

\section{Trajectory Controls}
\label{sec:app:controls}

\begin{figure*}[t]
  \centering
  \includegraphics[width=0.75\textwidth]{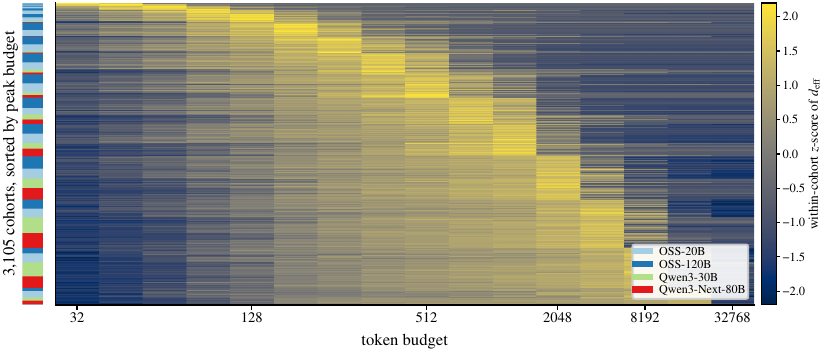}
  \caption{\textbf{The trajectory is a per-cohort fact, not an aggregation effect.}
  Each of the $\Ncohorts$ rows is one (model, question) cohort, standardised within
  the row and sorted by the budget at which it peaks; the left stripe gives the
  architecture. The bright band runs diagonally---the sort constructs the
  diagonal; the band's contrast, not its diagonality, carries the
  information---and the architecture stripe shows how the pooled peak combines
  architecture-specific clocks: OSS-20B cohorts concentrate at early peaks and
  Qwen3-Next-80B cohorts at late ones.}
  \label{fig:app:heatmap}
\end{figure*}

The low--high--low trajectory could in principle be produced by marginal activation
statistics, by a mixture of finished and running rollouts, or by aggregation. This
section expands the controls summarized in \Cref{sec:rq1};
\Cref{fig:app:heatmap} shows the per-cohort picture directly.

\paragraph{Expert identity carries the trajectory.}
Permuting expert identities per rollout and per layer---preserving per-rollout
activation counts, per-layer load, and sparsity exactly---reduces the prominent-peak
rate from $98.5\%$ to $0.0\%$ (the null rate is exactly $0.0\%$ in each of the ten
configurations), cuts median \emph{absolute} prominence from $15.92$ to $0.09$ (a
factor of $178$; median relative prominence falls $0.68\to0.00$), and
drives the spectrum to $\deff\!\approx\!n$ (median ratio $0.94$ to cohort size). An
interior argmax formally remains on $73.1\%$ of the resulting near-flat curves, so
the permutation destroys the peak's magnitude, not the existence of a maximum on a
noisy curve. The trajectory is therefore carried by cross-rollout expert
co-activation identities rather than aggregate routing mass.

\paragraph{Risk-set analysis preserves the arc and refines late-stage timing.}
The risk-set analysis isolates the contribution of still-generating rollouts at
large budgets, where the full graph otherwise combines running and completed
traces. Restricting the graph to the
\emph{risk set}---only rollouts still generating at $\bbudget$---leaves an interior
maximum in $99.8\%$ of cells, so frozen windows are not the sole source of the
non-monotonicity; but risk-set peak positions correlate only moderately with full-graph peaks
($\rho=+0.632$), so late-budget peak \emph{locations} partly reflect freezing. Once
every rollout has finished, $\deff$ is constant by construction (coefficient of
variation $0.0000$ at both the median and the $90$th percentile); post-completion
readings carry no dynamical content and are not interpreted.

\paragraph{Answer-token masking.}
Capping the trailing window at the boxed-answer anchor---so the formatted answer
characters never enter the histogram---leaves the trajectory statistically
unchanged: the prominent-peak rate is $98.7\%$ vs.\ $98.5\%$ ($\Delta$ $+0.002$
$[-0.005,+0.015]$, n.s.), and the peak frame is unmoved for $88.2\%$ of questions
($97.4\%$ within one frame); only the post-peak reconcentration becomes shallower
(collapse ratio $0.26\to0.40$), attributing roughly a fifth of the collapse depth
to the answer characters themselves. Masking the last $32$ pre-answer tokens
instead removes the arc ($98.5\%\to34.4\%$; per-configuration range
$4.6\%$--$70.6\%$), locating the signal in the commit region rather than in the
answer string. Mask distance is collinear with window staleness, so ``no answer
tokens'' and ``stale window'' are not fully separated.

\paragraph{The shape is temporal.}
Shuffling the order of each cohort's per-budget frames---preserving the marginal set
of graphs while destroying their sequence---drops the prominent-peak rate from
$98.5\%$ to $56.2\%$: the trajectory is a property of the time course, not of the
collection of frames.

\paragraph{Peak location is set by generation length.}
The peak's location tracks the model's own generation-length scale (rank correlation
$+0.94$ with the cohort's median completion budget), so a fixed token budget lands
at different trajectory positions in different models. Cross-model readings are
therefore taken in per-cohort peak units (\Cref{sec:app:protocol};
\Cref{tab:app:phasenorm}).

\section{Difficulty-Stratified Same-Answer Recoverability}
\label{sec:app:readability}

Same-answer AP is defined only for cohorts with $\ge2$ distinct answers; $38\%$ of
cohorts are unanimous and are excluded, leaving $\NdefinedAP$ (the confirmatory P2
audit additionally applies the audit family's $\ge$$10$-graded-rollout gate,
retaining $\NAPaudit$; \Cref{sec:app:audit}).
\Cref{tab:app:readability} gives the stratified recoverability curves, using the fixed
majority-share cutpoints $[0.60,0.875]$ plus the most-contested slice; the dynamic
coupling of AP \emph{changes} to the spectral increments $\Delta m$ and $\Delta r$
is analyzed in \Cref{sec:rq3}.

\begin{table*}[t]
  \centering
  \caption{Same-answer AP by budget and difficulty stratum (defined-AP cohorts). The
  column grid omits $\bbudget{=}4096$ for space; the pooled defined-AP there is
  $0.673$ (between $\bbudget{=}2048$'s $0.657$ and $\bbudget{=}8192$'s $0.688$). The
  last column is the same-answer AP of the commit-window graph $\Wc$ itself
  ($0.719$ pooled); b32768 is the largest prefix budget. Absolute levels are not
  comparable across strata (each stratum has its own same-answer prevalence, hence
  its own random-AP baseline).}
  \label{tab:app:readability}
  \small
  \begin{tabular}{lccccccc}
    \toprule
    Stratum & b32 & b128 & b512 & b2048 & b8192 & b32768 & $\Wc$ \\
    \midrule
    easy ($\ge0.875$)     & $0.901$ & $0.902$ & $0.909$ & $0.917$ & $0.930$ & $0.940$ & $0.940$ \\
    mid ($[0.60,0.875)$) & $0.605$ & $0.612$ & $0.625$ & $0.653$ & $0.692$ & $0.725$ & $0.726$ \\
    hard ($<0.60$)        & $0.317$ & $0.321$ & $0.337$ & $0.394$ & $0.433$ & $0.480$ & $0.484$ \\
    contested ($<0.5$)    & $0.237$ & $0.239$ & $0.247$ & $0.315$ & $0.357$ & $0.413$ & $0.418$ \\
    \midrule
    all defined-AP        & $0.611$ & $0.614$ & $0.627$ & $0.657$ & $0.688$ & $0.717$ & $0.719$ \\
    \bottomrule
  \end{tabular}
\end{table*}

\section{Spectral Anatomy: Details, Controls, and Attribution Tests}
\label{sec:app:anatomy}

\begin{figure}[t]
  \centering
  \includegraphics[width=0.55\textwidth]{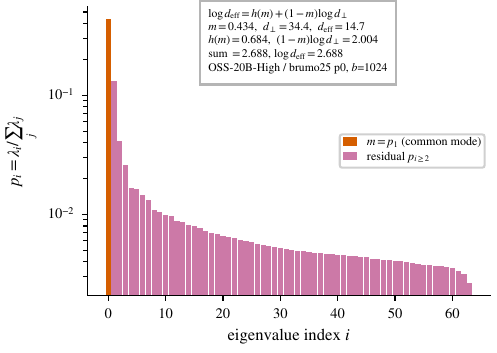}
  \caption{\textbf{The exact identity on one real, typical eigenspectrum.} A
  normalised eigenspectrum (log scale) with the common-mode mass $m=p_1$
  highlighted; the annotation evaluates $\log\deff=h(m)+(1-m)\log\dperp$ term by
  term on this frame, and the two sides agree to the printed precision. The
  example is the committed per-shard spectrum whose $m$ is closest to the median
  of those examples---typical rather than cherry-picked.}
  \label{fig:app:identity}
\end{figure}

\begin{figure*}[t]
  \centering
  \includegraphics[width=0.75\textwidth]{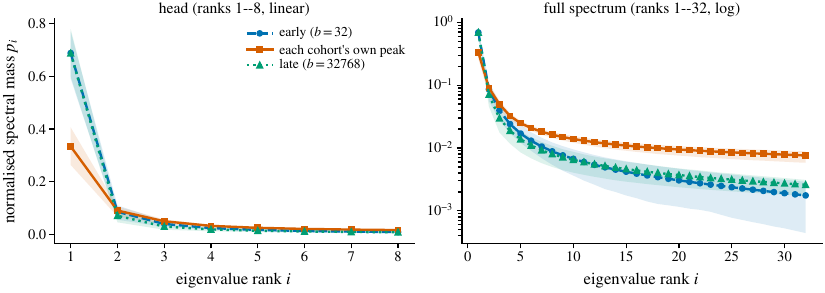}
  \caption{\textbf{The trajectory is a redistribution of spectral mass.} Mean
  normalised eigenspectrum over $\Ncomplete$ cohorts (band: IQR), at the first budget,
  at each cohort's \emph{own} peak (peak-aligned, since family peak budgets differ
  several-fold; \Cref{tab:app:calipers}), and at the last budget. The leading direction holds
  $p_1{=}0.689$ early, $0.334$ at the peak and $0.688$ late, while the tail beyond
  rank $8$ moves the opposite way ($0.084/0.238/0.096$). The low--high--low form in
  $\deff$ is exactly this shift of mass out of and back into the leading
  eigendirections. Note that the two ends are spectrally alike, not merely equal in
  $\deff$: $p_1$ differs by $0.001$ between them. Cohorts with fewer than $32$
  graded rollouts are excluded so that every curve is averaged over the same
  eigenvalue ranks.}
  \label{fig:app:spectra}
\end{figure*}

This section backs \Cref{sec:rq2}. The identity of \Cref{eq:decomp} is evaluable
wherever $\dperp$ is defined ($m<1$): $\Nident$ of the $\Nframes$ frames ($16$
budgets $\times$ $\Ncohorts$ cohorts; the $131$ frames at $m\approx1$ have undefined
$\dperp$, and $\Ncomplete$ cells have all $16$ frames defined). Its maximum
per-frame residual is $5.7\times10^{-14}$; per-segment attribution residuals are
${\le}1.3\times10^{-15}$.
\Cref{fig:app:identity} evaluates the identity term by term on one real
eigenspectrum; \Cref{fig:app:spectra} shows the underlying spectra directly;
\Cref{tab:app:attrib} attributes each trajectory segment's motion to the three
channels; \Cref{tab:app:cvr2} fits the residual axis on the common mode by subset;
\Cref{tab:app:anatomynull} reports the experiment's nulls.

\begin{table*}[t]
  \centering
  \caption{Attribution of trajectory motion by the midpoint split of
  \Cref{eq:decomp}: share of $|\Delta\log\deff|$ carried by the mixing term $h(m)$,
  the common-mode term, and the residual term, per segment (two-way clustered $95\%$
  CIs; all common-mode and residual shares are positive in $\ge9/10$ configurations;
  mixing shares are n.s.\ with MDE$_{80}$ $0.04$--$0.08$). ``Live'' restricts the
  segment to still-generating rollouts. Shares are pooled ratio-of-sums over each
  arm's full cell set; the pre-registered H2 contrast of \Cref{sec:app:anatomy} is
  the same functional on the $2{,}604$ cells matched across both halves (excluding
  $395$ late-half segments whose $\Delta\log\deff$ is exactly zero under frozen
  windows), where the early share is $0.317$---so the contrast is $+0.158$, not the
  difference of the displayed rows.}
  \label{tab:app:attrib}
  \small
  \setlength{\tabcolsep}{4pt}
  \begin{tabular}{lrrccc}
    \toprule
    Segment & $n$ & $\Delta\log\deff$ & mixing $h$ & common mode & residual \\
    \midrule
    early half ($32\!\to\!1024$)   & $2999$ & $+1.136$ & $+0.026$ & $+0.631$ $[+.537,+.719]$ & $+0.342$ $[+.271,+.429]$ \\
    late half ($1024\!\to\!32768$) & $2999$ & $-0.920$ & $+0.017$ & $+0.824$ $[+.784,+.868]$ & $+0.158$ $[+.117,+.189]$ \\
    late half, live only           & $2256$ & $-0.717$ & $+0.005$ & $+0.776$ $[+.707,+.845]$ & $+0.219$ $[+.170,+.267]$ \\
    rise (peak-aligned)            & $2931$ & $+1.605$ & $+0.020$ & $+0.691$ $[+.611,+.772]$ & $+0.289$ $[+.212,+.354]$ \\
    fall (peak-aligned)            & $2931$ & $-1.400$ & $+0.012$ & $+0.853$ $[+.822,+.885]$ & $+0.135$ $[+.092,+.174]$ \\
    fall, live only                & $2663$ & $-1.075$ & $-0.002$ & $+0.836$ $[+.799,+.874]$ & $+0.166$ $[+.112,+.216]$ \\
    \bottomrule
  \end{tabular}
\end{table*}

\paragraph{Pre-registered hypotheses.}
Three hypotheses formed the Bonferroni family ($\alpha=0.05/3$, $98.33\%$
intervals). \emph{H1} (common-mode share exceeds residual share): early
$+0.289$, $95\%$ $[+0.110,+0.444]$, $98.33\%$ $[+0.044,+0.478]$, $9/10$; late
$+0.666$ $[+0.612,+0.731]$, $98.33\%$ $[+0.599,+0.756]$, $10/10$---both supported.
\emph{H2} (residual share larger early than late; matched-cell contrast,
$n=2{,}604$): $+0.158$ $[+0.080,+0.232]$,
$98.33\%$ $[+0.060,+0.251]$, $10/10$---supported; with the late half restricted to
live rollouts ($n=2{,}256$) the contrast attenuates $45\%$ to $+0.087$
$[+0.011,+0.158]$, $8/10$,
and across seven bootstrap seeds the lower bound ranges $+0.009$ to $+0.016$: the
contrast remains positive but attenuated under the live-rollout restriction, with
a modestly seed-sensitive lower confidence bound. \emph{H3} (residual share enriched near the peak): rise
$-0.020$ $[-0.133,+0.075]$, n.s.\ (MDE$_{80}$ $0.146$); fall $+0.269$
$[+0.213,+0.318]$, nominally positive---but a matched permuted-peak placebo
reproduces $94\%$ of it (excess $+0.016$; $+0.006$ to $+0.021$ across estimators)
and cross-problem pseudo-cohorts reproduce roughly two thirds, so under the
pre-registered placebo rule we interpret the contrast as a budget-position effect
rather than peak-specific enrichment.

\paragraph{Placebos and cut points.}
Permuting each problem's peak position across problems reproduces the attribution
split almost exactly (placebo common-mode share $0.611$ rise / $0.831$ fall against
true $0.691/0.853$, i.e.\ $88\%/97\%$), and a fixed-cut-point sweep with no peak
anywhere gives a similar split at every cut: the decomposition describes the budget
axis, not the peak.

\paragraph{Headroom normalization.}
Normalised by each factor's available headroom, early-half motion is
indistinguishable between the two factors ($m$ $+0.282$ $[+0.214,+0.359]$, $r$
$+0.226$ $[+0.139,+0.333]$; difference $+0.056$ $[-0.014,+0.116]$, n.s.,
MDE$_{80}$ $0.093$); in the late half the common mode does move faster
($+0.140$ $[+0.112,+0.167]$). The nats-scale dominance of the common-mode term thus
reflects a $5.89\times$ multiplier ratio in \Cref{eq:decomp}, not uniformly faster
motion of $m$.

\paragraph{Architecture-aware family contrasts.}
Because family co-varies with routing architecture, these contrasts are
interpreted descriptively rather than causally. Significant residual-share differences
concentrate on the descending branch (rise \textsf{Instruct}$-$\textsf{OSS}
$+0.154$ $[+0.065,+0.273]$; fall \textsf{Thinking}$-$\textsf{OSS} $+0.097$
$[+0.043,+0.156]$; late \textsf{Thinking}$-$\textsf{OSS} $+0.088$
$[+0.045,+0.155]$; fall and late \textsf{Instruct}$-$\textsf{Thinking} $-0.051$ and
$-0.075$, both negative). The present sample does not resolve early-segment family
differences (MDE$_{80}$ $0.24$--$1.77$), so these contrasts are left
uninterpreted. Shoelace tests do not detect family-level hysteresis in the $(m,r)$
plane under the tested variants (e.g.\ closed loop $-0.011$ $[-0.070,+0.039]$,
$5/10$).

\begin{table}[t]
  \centering
  \caption{Grouped cross-validated fit of $r=\log\dperp$ on $m$ (folds grouped by
  question; $\Ncvframes$ frames from the $\Ncomplete$ complete cells). The residual
  axis is only partially predictable from the common mode.}
  \label{tab:app:cvr2}
  \small
  \setlength{\tabcolsep}{4pt}
  \begin{tabular}{lrcc}
    \toprule
    Subset & frames & CV $R^2$ & Spearman \\
    \midrule
    all                      & $47984$ & $0.369$ & $-0.639$ \\
    \textsf{OSS}             & $29808$ & $0.295$ & $-0.530$ \\
    \textsf{Qwen-Instruct}   & $9824$  & $0.533$ & $-0.764$ \\
    \textsf{Qwen-Thinking}   & $8352$  & $0.579$ & $-0.767$ \\
    all, rising branch       & $31109$ & $0.545$ & $-0.724$ \\
    all, falling branch      & $18718$ & $0.303$ & $-0.345$ \\
    \textsf{OSS}, falling    & $13067$ & $0.251$ & $-0.125$ \\
    \bottomrule
  \end{tabular}
\end{table}

\begin{table}[t]
  \centering
  \caption{Nulls for the anatomy experiment, on its own $6$-shard pool and
  prominence gate (not comparable to the full-corpus gate;
  \Cref{sec:app:protocol}). The expert-identity permutation drives the residual
  spectrum toward its near-isotropic ceiling ($\dperp\!\to\!N{-}1$); cross-problem
  pseudo-cohorts of unrelated rollouts retain the coarse arc.}
  \label{tab:app:anatomynull}
  \footnotesize
  \setlength{\tabcolsep}{1.5pt}
  \begin{tabular}{lccccc}
    \toprule
    Variant & $n$ & med.\ $m$ & med.\ $\dperp$ & med.\ $\deff$ & prom. \\
    \midrule
    real ($N{=}64$)            & $204$ & $0.450$ & $38.9$ & $14.4$ & $97.5\%$ \\
    id.\ permutation           & $204$ & $0.082$ & $62.9$ & $59.5$ & $3.4\%$ \\
    pseudo-cohorts ($N{=}32$)  & $264$ & $0.321$ & $25.1$ & $16.8$ & $97.0\%$ \\
    \bottomrule
  \end{tabular}
\end{table}

\section{Behavioral Localization: Preregistered Plan and Predictive Ladders}
\label{sec:app:audit}

\paragraph{The frozen plan.}
The audit's endpoints, decision rule, and verdict sentences were fixed before the
analyses ran. For each target $Y$ the predictive specification ladder is
$Y\!\sim\!\mathrm{controls}$; ${+}\,f(\dens)$; ${+}\,f(m)$; ${+}\,f(m)+g(r)$;
${+}\,f(\log\deff)$, with $f,g$ cubic B-spline bases ($5$ knots, fit on train
folds) and controls $=$ model$\times$dataset fixed effects $+$ log cohort size $+$
two cohort-accuracy covariates (an indicator for majority-incorrect cohorts and
terminal cohort accuracy; answer-derived, which if anything makes the residual
null conservative), evaluated by $5$-fold GroupKFold with groups $=$ question. The two
headline quantities are $\Delta R^2_{r\mid m}=R^2_{m+r}-R^2_m$ (residual-spectrum
value) and $\Delta R^2_{\mathrm{split}}=R^2_{m+r}-R^2_{\deff}$
(the split-spectrum-vs-scalar out-of-fold difference). These are out-of-fold
predictive comparisons rather than formally nested models (only $m\to(m,r)$ is
nested). Primaries (Bonferroni family of three,
$\alpha=0.0167$): \emph{P1} answer effective count $\log\dans$ against commit-graph
state; \emph{P2} terminal AP lift on defined-AP cohorts; \emph{P3} the paired effort
contrast on peak height. Correctness was pre-registered as an uncorrected
\emph{boundary} endpoint. The decision tree committed three verdict sentences
verbatim, and the frozen rule selected Outcome C; for neutral presentation the
branches are labelled A/B/C, with the verbatim verdicts and the selection rule
unchanged.
\textbf{Outcome A:} ``Effective rank is common-mode dominated, but its residual
spectrum carries independent information about semantic multiplicity and
routing--answer organization---not correctness.''
\textbf{Outcome B:} ``Residual spectral dimensionality reflects computation style
rather than output semantics.''
\textbf{Outcome C (selected):} ``In
these cohorts, routing effective rank is primarily a
smooth full-spectrum reparameterization of global routing agreement; the residual
axis varies, but we find no independent operational semantics for it.''
Outcome C constrains claims about incremental residual semantics on the tested
answer-space targets; it does not alter the exact spectral decomposition, the
residual axis's structural independence, or the effort--timing effects.

\paragraph{Behavioral-localization layer.}
The confirmatory verdict above is unchanged, and P3's registered primary---the
paired effort contrast on peak height---varies across architectures with a pooled
interval spanning zero
($+0.104$ $[-0.079,+0.232]$; \Cref{tab:app:effort}). On top of it
we report secondary trajectory endpoints that localize the positive effects rather
than alter the primary family: peak time, peak-normalized occupancy
$\mathrm{Occ}_d$, width above $0.8d_{\max}$, and integrated common-mode/residual
mass. Peak time, occupancy, and the integrated masses are passthroughs of the
confirmatory statistics; the width and alignment readouts were computed under
decision rules fixed before this layer ran. The width rule
licensed ``broadens'' only if all four architectures agreed and the pooled interval
excluded zero; both conditions were met. The pooled rows are descriptive aggregates
over four architecture clusters. A separate peak-alignment analysis reveals
architecture-specific shape changes beyond a rigid temporal translation.

\begin{table*}[t]
  \centering
  \caption{Semantic-audit ladders: out-of-fold $R^2$ under GroupKFold by question.
  $\Delta R^2_{r\mid m}$ carries two-way clustered $95\%$ CIs; the Bonferroni
  $98.33\%$ intervals for the two regression primaries are P1 $[-0.011,+0.052]$ and
  P2 $[-0.102,+0.036]$---both span zero. Density and common-mode mass provide
  nearly identical predictive value on these endpoints ($\Delta R^2$: $-0.0005$ on
  P1, $+0.0036$ on P2), localizing the observable association to the global
  agreement channel; the spectral formulation contributes exact channel
  attribution and an independent residual coordinate rather than scalar
  predictive lift.}
  \label{tab:app:ladder}
  \small
  \setlength{\tabcolsep}{3.5pt}
  \begin{tabular}{lrccccccc}
    \toprule
    Target & $n$ & controls & ${+}\dens$ & ${+}m$ & ${+}(m,r)$ & ${+}\log\deff$ &
    $\Delta R^2_{r\mid m}$ [$95\%$] & $\Delta R^2_{\mathrm{split}}$ \\
    \midrule
    $\log\dans$ (commit, P1)     & $\Naudit$ & $0.610$ & $0.631$ & $0.631$ & $0.648$ & $0.641$ & $+0.018$ $[-.004,+.041]$ & $+0.007$ \\
    $\log\dans$ (terminal, sens.)& $\Naudit$ & $0.610$ & $0.632$ & $0.632$ & $0.649$ & $0.642$ & $+0.017$ $[-.018,+.039]$ & $+0.007$ \\
    AP lift (terminal, P2)       & $\NAPaudit$ & $0.216$ & $0.260$ & $0.264$ & $0.267$ & $0.268$ & $+0.003$ $[-.051,+.023]$ & $-0.002$ \\
    difficulty (secondary)       & $3096$ & $0.564$ & $0.586$ & $0.586$ & $0.607$ & $0.595$ & $+0.022$ $[-.001,+.050]$ & $+0.012$ \\
    \bottomrule
  \end{tabular}
\end{table*}

\begin{figure*}[t]
  \centering
  \includegraphics[width=0.75\textwidth]{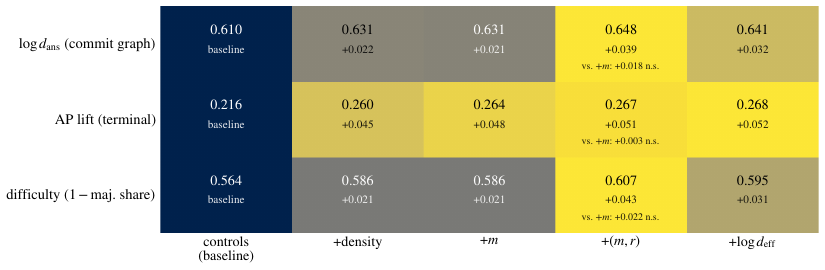}
  \caption{\textbf{Out-of-sample model ladders for the semantic endpoints.}
  Grouped-CV $R^2$ (upper number) and its increment over controls (lower number).
  The predictive ladder localizes answer-space coupling to the global agreement
  channel: density and common-mode mass provide nearly identical predictive value
  on these targets, while the residual axis remains the distinct structural
  coordinate of RQ2 (increments bounded in
  \Cref{fig:app:residualscope}). The scalar $\log\deff$ lies between the
  agreement-only and split-spectrum models.}
  \label{fig:app:ladder}
\end{figure*}

\begin{figure}[t]
  \centering
  \includegraphics[width=\columnwidth]{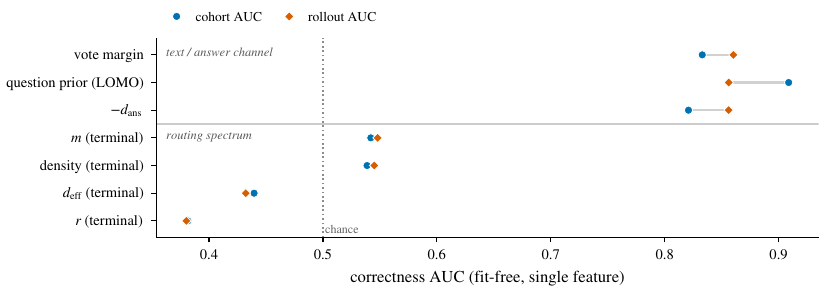}
  \caption{\textbf{Correctness selection favors answer-side statistics.}
  Fit-free single-feature AUCs for cohort majority correctness (circles) and
  rollout-level correctness (diamonds); dotted line: chance; two-way clustered
  $95\%$ CIs ($B{=}500$, uncorrected). Text-side baselines sit
  at $0.82$--$0.91$ (the $0.91$ question prior uses the other nine configurations'
  outcomes), while the best routing feature reaches $0.542$, separating answer
  selection from the structural-monitoring role studied in the main paper.}
  \label{fig:app:correctness}
\end{figure}

\Cref{tab:app:ladder} gives the full ladders (\Cref{fig:app:ladder} renders them
as a heatmap); per-architecture effort contrasts are in \Cref{tab:app:effort}, and
the correctness boundary in \Cref{tab:app:correctness} and
\Cref{fig:app:correctness}.

\paragraph{Conditioning audit.}
Every subset gate reports kept and dropped counts to make the conditioning step
explicit. P1 keeps $\Naudit/\Ncohorts$ (kept mean $m$ $0.688$ vs.\
dropped $0.628$). P2 keeps $\NAPaudit/\Ncohorts$: $\Nunanimous$ unanimous cohorts
have undefined AP (leaving $\NdefinedAP$), and a further $6$ are removed by the
audit family's shared $\ge$$10$-graded-rollout gate (all
\textsf{OSS-20B-High}$\times$GPQA cohorts with $5$--$9$ gradeable answers); the
dropped set has majority share $0.998$ vs.\ $0.715$ kept---so the alignment verdict
speaks only for contested, sufficiently graded cohorts.

\paragraph{Dynamic recoverability localization.}
Regressing per-step changes in AP lift on the spectral increments with budget-step
fixed effects ($\Nsteps$ steps from $318$ questions $\times$ $10$ configurations)
gives $\mathrm{coef}(\Delta m)=+0.522$
$[+0.430,+0.609]$ and $\mathrm{coef}(\Delta r)=-0.047$
$[-0.092,-0.016]$ (two-way question$\times$configuration clustered, $B{=}500$; a
supplementary analysis outside the Bonferroni family); a within-question time
shuffle sits at
${\approx}+0.002/{-}0.000$. Thus the positive recoverability association is localized
to increasing common-mode mass; the residual partial is small and opposite-signed.
Standardized, a $1$-SD step in common-mode mass moves recoverability roughly four
times as much as a $1$-SD step in residual dimensionality ($+0.048$
$[+0.040,+0.057]$ vs.\ $-0.011$ $[-0.023,-0.004]$ per SD); re-clustering on
question $\times$ base architecture (four clusters) widens both intervals by
$15$--$40\%$ without changing either sign ($\beta_{\Delta r}$
$[-0.111,-0.015]$).

\begin{table*}[t]
  \centering
  \caption{Effort localization, paired within architecture (higher minus lower
  tier, matched by question; question-clustered $95\%$ CIs; pooled row clustered by
  question$\times$architecture, $4$ architecture clusters). $\Delta\log d_{\max}$
  (peak height) is the registered primary endpoint of this contrast---it varies by
  architecture, with a pooled interval spanning zero; the remaining columns are
  secondary endpoints.
  $\mathrm{Occ}_d=\int d_{\mathrm{eff}}(\tau)/d_{\max}\,\mathrm{d}\tau$ with $\tau$
  the normalized budget-grid index; $w_{0.8}$ is the octave width of the contiguous
  $d_{\mathrm{eff}}\ge0.8d_{\max}$ regime. The consistent effects are later maxima,
  greater peak-normalized occupancy, a wider high-rank interval, and
  weaker integrated common-mode mass. Residual-spectrum area has no common shift.}
  \label{tab:app:effort}
  \scriptsize
  \setlength{\tabcolsep}{1.25pt}
  \resizebox{\textwidth}{!}{%
  \begin{tabular}{lrcccccc}
    \toprule
    Architecture & $n$ & $\Delta\log_2 t_{\max}$ & $\Delta\log d_{\max}$ &
    $\Delta\mathrm{Occ}_d$ &
    $\Delta w_{0.8}$ & $\Delta\mathrm{AUC}_m$ & $\Delta\mathrm{AUC}_r$ \\
    \midrule
    OSS-20B       & $282$ & $+3.86\,[3.66,4.03]$ & $+.181\,[.141,.217]$ & $+.107\,[.097,.117]$ & $+1.47\,[1.28,1.63]$ & $-.113\,[-.119,-.106]$ & $-.144\,[-.195,-.093]$ \\
    OSS-120B      & $309$ & $+2.97\,[2.80,3.14]$ & $+.234\,[.189,.285]$ & $+.172\,[.161,.183]$ & $+2.02\,[1.84,2.20]$ & $-.156\,[-.163,-.150]$ & $+.104\,[.064,.146]$ \\
    Qwen3-30B     & $310$ & $+1.70\,[1.56,1.83]$ & $+.093\,[.058,.124]$ & $+.040\,[.030,.052]$ & $+.47\,[.29,.65]$ & $-.038\,[-.045,-.031]$ & $+.079\,[.040,.117]$ \\
    Qwen3-Next-80B& $296$ & $+1.90\,[1.82,2.00]$ & $-.093\,[-.130,-.052]$ & $+.087\,[.078,.096]$ & $+1.22\,[1.07,1.36]$ & $-.041\,[-.048,-.033]$ & $-.039\,[-.080,.006]$ \\
    \midrule
    Pooled        & ---   & $+2.59\,[1.72,3.83]$ & $+.104\,[-.079,.232]$ & $+.102\,[.045,.168]$ & $+1.29\,[.55,1.99]$ & $-.087\,[-.154,-.036]$ & $+.004\,[-.127,.111]$ \\
    \bottomrule
  \end{tabular}}
\end{table*}

\begin{figure*}[t]
  \centering
  \includegraphics[width=0.80\textwidth]{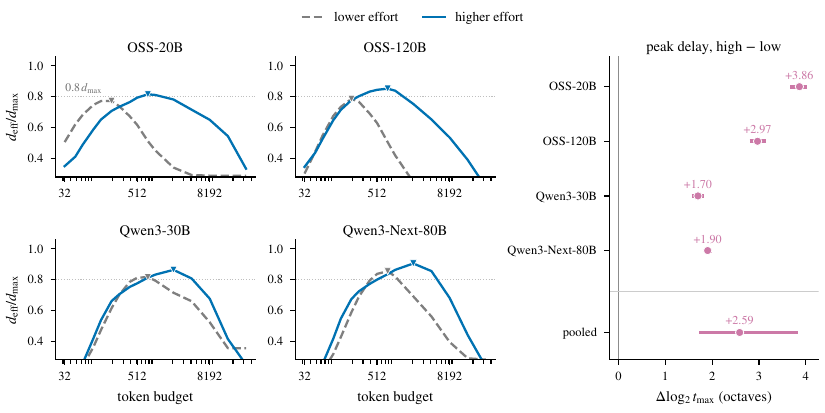}
  \caption{\textbf{Higher effort moves the effective-rank maximum later in every
  architecture.} Peak-normalized trajectories for paired lower- and higher-effort
  configurations, with paired peak-delay intervals on the right. The timing effect is
  uniform; peak height is architecture-dependent.}
  \label{fig:app:efforttraj}
\end{figure*}

\paragraph{Peak alignment does not collapse the effort contrast.}
After aligning each cohort by its own maximum, the lower/higher-effort trajectory gap
shrinks by at least half only for OSS-20B; the aligned/raw gap ratios are
$0.32/0.74/2.24/1.62$ across the four architectures. The effort effect is therefore
not a rigid time translation of one shared curve (\Cref{fig:app:efforttraj} shows
the unaligned trajectories). Edge censoring is negligible
($1.8\%$); excluding censored pairs leaves the width contrast unchanged
($+1.26$ $[+0.57,+1.97]$ octaves).

\begin{table}[t]
  \centering
  \caption{Correctness comparison: fit-free single-feature AUCs (cohorts with
  $\ge10$ graded rollouts). We retain the pre-registered one-sided reading and do
  not reinterpret the coefficient sign post hoc; answer-side baselines remain
  stronger than standalone routing features for correctness selection.}
  \label{tab:app:correctness}
  \small
  \setlength{\tabcolsep}{4pt}
  \begin{tabular}{lcc}
    \toprule
    Feature & cohort AUC [$95\%$] & rollout AUC \\
    \midrule
    vote margin              & $0.833$ $[.790,.869]$ & $0.860$ \\
    $\dans$ (neg.)           & $0.821$ $[.777,.859]$ & $0.856$ \\
    question prior (LOMO)    & $0.909$ $[.872,.953]$ & $0.856$ \\
    \midrule
    $m$ (terminal)           & $0.542$ $[.472,.658]$ & $0.548$ \\
    density (terminal)       & $0.539$ $[.469,.660]$ & $0.545$ \\
    $\deff$ (terminal)       & $0.440$ $[.324,.509]$ & $0.432$ \\
    $r$ (terminal)           & $0.381$ $[.321,.478]$ & $0.380$ \\
    \bottomrule
  \end{tabular}
\end{table}

\begin{figure}[t]
  \centering
  \includegraphics[width=\columnwidth]{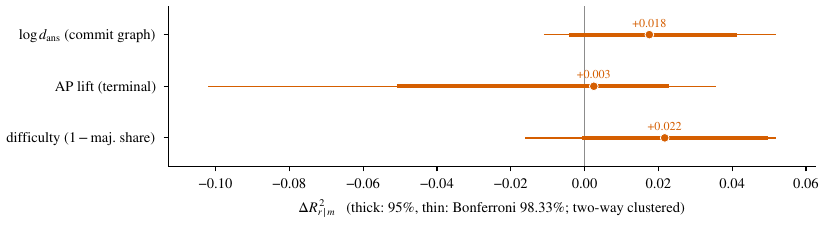}
  \caption{\textbf{Residual \emph{terminal incremental} associations are bounded
  on the tested targets.} Incremental out-of-fold $R^2$ from adding residual dimensionality
  after common-mode mass. Thick whiskers are $95\%$ and thin whiskers Bonferroni
  $98.33\%$ two-way clustered intervals. The intervals constrain independent
  residual contributions on the tested targets ($95\%$ upper bounds: $+0.04$ for
  answer effective count, $+0.02$ for AP lift) while remaining compatible with small
  effects.}
  \label{fig:app:residualscope}
\end{figure}

\section{Subcohort Fidelity and Absolute Scale}
\label{sec:app:subcohort}

\begin{table}[t]
  \centering
  \caption{Rank fidelity of subsampled $\deff$: median over $50$ (model, dataset)
  shards of the Spearman correlation, across a shard's cohorts, between $\deff$ on
  a random $n$-rollout subcohort ($20$ draws) and the full $64$-rollout graph;
  two-way (dataset $\times$ model) clustered $95\%$ CIs.}
  \label{tab:app:subcohort}
  \small
  \setlength{\tabcolsep}{4pt}
  \begin{tabular}{rcc}
    \toprule
    $n$ & terminal frame & peak frame \\
    \midrule
    $2$  & $0.517$ $[.430,.588]$ & $0.200$ $[.158,.265]$ \\
    $4$  & $0.738$ $[.654,.763]$ & $0.411$ $[.315,.485]$ \\
    $8$  & $0.850$ $[.788,.880]$ & $0.611$ $[.493,.692]$ \\
    $16$ & $0.926$ $[.889,.942]$ & $0.791$ $[.721,.838]$ \\
    $32$ & $0.971$ $[.957,.976]$ & $0.916$ $[.890,.945]$ \\
    \bottomrule
  \end{tabular}
\end{table}

\begin{figure}[t]
  \centering
  \includegraphics[width=0.50\textwidth]{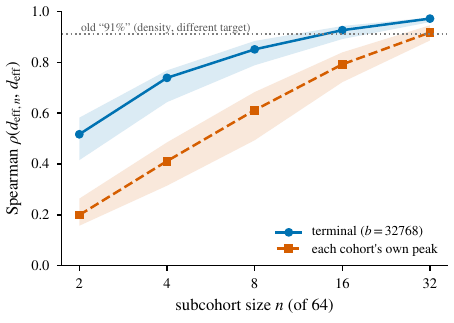}
  \caption{\textbf{Rank fidelity of subsampled $\deff$} (\Cref{tab:app:subcohort}):
  terminal-frame ordering is reliable from $n\!\approx\!8$ ($\rho=0.85$), while
  peak-frame fidelity improves steadily from $0.20$ at $n{=}2$ to $0.92$ at
  $n{=}32$, reflecting the greater spectral spread near the trajectory maximum.
  The dotted line at $0.91$ is the fidelity of per-rollout \emph{density}
  against a different (same-answer recoverability) target; it does not describe $\deff$.}
  \label{fig:app:subcohort}
\end{figure}

Small-cohort estimates of $\deff$ order cohorts usably at the terminal frame from
$n\!\approx\!8$, but the peak frame---where the spectrum is most spread---requires
far larger subcohorts (\Cref{tab:app:subcohort}, \Cref{fig:app:subcohort}).

\paragraph{Absolute scale.}
An effective rank computed on $n$ rollouts is bounded by $n$, so small-cohort
readings are compressed by construction and can only be compared across cohort
sizes after an inversion; for rank-based uses the compression is close to a
monotone rescaling. Where an absolute value is wanted, a closed form is available
for the \emph{order-two} effective rank, which depends on the graph only through
its mean squared off-diagonal entry; because that quantity is a pairwise mean, a
subsample estimates it without bias and the full-cohort value follows with no
fitted calibration: median error $4.4\%$ at $n{=}8$, versus $7.5\%$ for a
per-family fitted inversion. Within the Hill family~\citep{hill1973diversity} the
order-two member is the one that admits this; the order-one member underlying
$\deff$ does not. The same construction removes the cohort-size dependence that
otherwise contaminates size-varying readouts.

\section{Frozen-Coordinate Transfer to LiveCodeBench-v5}
\label{sec:app:transfer}

Every quantity in the paper was developed on math and science benchmarks. To test
whether the trajectory readouts are properties of that domain or of the process, we
froze every dial---window length, subcohort size $n{=}8$, peak-detector thresholds,
all family constants---and ran the readouts once on LiveCodeBench-v5, which had
been excluded from every analysis decision ($1{,}582$ problems, the same ten
configurations). The gross trajectory statistics transfer within noise of their math
anchors: the share of cohorts with a prominent interior maximum is $97.7\%$ (math:
$98.5\%$), a frozen running-maximum peak detector fires on $98.3\%$ of cohorts at
precision $0.972$ (math: $0.92$--$0.95$), and the recoverability gain transfers with
the same sign at smaller magnitude. \textsf{Thinking} transfers best (peak share
$100\%$).

Transfer sharpens two scope conditions. Cross-problem rank fidelity at $n{=}8$
($\bbudget{=}4096$) is $0.881$, against a matched within-configuration math value
of $0.866$; the corpus-pooled variant of the same statistic reads ${\approx}0.98$
only because between-configuration level differences dominate its ranking, and is
not a valid anchor here (this readout is also distinct from the terminal/peak
fidelities of \Cref{tab:app:subcohort}, which use the full $64$-rollout graph and
single draws). The domain-sensitive readout is the \emph{text} channel:
string-match vote margins do not transfer to code, reading contested cohorts at
AUC $0.433$ against
${\approx}0.83$ on math, because functionally similar programs rarely share
identical strings; the routing-side correctness readout is unchanged ($0.407$,
inside its null band). The division of labour---text statistics for answer selection---%
presumes an answer space in which string equality is meaningful; execution-based
verification is the right substitute where it is not.

\section{Representation Robustness: Binary Expert-Set Jaccard}
\label{sec:app:binary}

All main results use the weighted-Jaccard (gate-weight) graph. Replacing it with a
\emph{binary} expert-set Jaccard---which experts fired, ignoring gate
weights---leaves every soft/geometric result unchanged or slightly better, on the
same $\Ncohorts$ cohorts (the two pipelines retain identical problem sets, so the
comparison is matched by construction). The family ordering is preserved ($\deff$
$7.5/14.9/19.6$ vs.\ weighted $9.2/18.8/25.0$; only the absolute scale compresses);
same-answer AP improves by $+0.006$ to $+0.008$. The partition and
correctness-selection conclusions are weight-independent and unchanged. More
importantly, binary expert-activation sets preserve the routing fingerprint and
slightly improve same-answer AP: the method operates from sparse
expert-activation indicators alone, without router gate magnitudes.

\section{Spectral Validity of the Routing Graph}
\label{sec:app:psd}

Empirically, the symmetrized weighted-Jaccard (Ruzicka) matrices are positive
semi-definite to numerical precision here. Across all $\Ngraphs$ cohort graphs
($17\times\Ncohorts$: every model, dataset, and budget, plus the commit window) the
minimum eigenvalue of the symmetrized $\Wmat$ is $\ge-5.8\times10^{-14}$
($\ge-4.6\times10^{-14}$ on the boxed-rollout submatrices the spectra are computed
on); the fraction of graphs with a minimum eigenvalue below $-10^{-12}$ is
$0.00\%$. The clip to zero
used before the spectral entropy (\Cref{eq:deff}) therefore removes only
floating-point noise.

\begin{table*}[t]
  \centering
  \caption{Sample accounting. All analyses draw from the same fixed activation
  cache; the retained count differs only by an explicit per-analysis filter.}
  \label{tab:app:accounting}
  \small
  \resizebox{\textwidth}{!}{%
  \begin{tabular}{llll}
    \toprule
    Analysis & Retained & Unit & Filter \\
    \midrule
    Pooled trajectory (\Cref{fig:trajectory}) & $\Ncohorts$ & cohorts/cell & none (frozen windows retained; all $16$ frames finite) \\
    Cohort corpus (\Cref{sec:exp:setup}) & $\Ncohorts$ & cohorts & 5 math/sci, 50 shards \\
    Binary-Jaccard ablation (\Cref{sec:app:binary}) & $\Ncohorts$ & cohorts & same corpus, binary graph \\
    Spectral anatomy (\Cref{sec:app:anatomy}) & $\Nident$ / $\Ncomplete$ & frames / cells & $\dperp$ defined ($m{<}1$) / all $16$ frames \\
    Audit P1 / difficulty (\Cref{tab:app:ladder}) & $\Naudit$ & cohorts & $\ge10$ graded rollouts \\
    Audit P2, AP lift (\Cref{tab:app:ladder}) & $\NAPaudit$ & cohorts & defined AP ($\ge2$ answers) and $\ge10$ graded rollouts \\
    Dynamic AP regression (\Cref{sec:app:audit}) & $\Nsteps$ & budget steps & defined-AP increments \\
    Subcohort fidelity (\Cref{tab:app:subcohort}) & $\Ncohorts$ & cohorts & 50 shards $\times$ 20 draws \\
    Defined-AP recoverability (\Cref{tab:app:readability}) & $\NdefinedAP$ & cohorts & $\ge2$ distinct answers (no audit gate) \\
    Token$\times$acc Pareto (\Cref{sec:app:frontier}) & $2{,}996$ & problems & exact rollout lengths \\
    Spectral validity (\Cref{sec:app:psd}) & $\Ngraphs$ & graphs & every model$\times$data$\times$budget + commit \\
    \bottomrule
  \end{tabular}}
\end{table*}

\section{Operational Scope: Structural Monitoring versus Answer Selection}
\label{sec:app:boundary}

We additionally tested whether the global spectral summaries that diagnose cohort
organization can support answer selection. Across fit-free correctness AUCs,
partition tests, vote reweighting, and token--accuracy frontier analyses,
answer-side signals remain stronger for selection. These results separate two
complementary roles: routing spectra provide a label-free view of internal cohort
geometry, while margins, execution signals, and verifiers remain the appropriate
tools for answer selection.

\subsection{Structural Analysis: Routing Geometry Is Diffuse Rather than
Partition-Like}
\label{sec:app:boundary:struct}

\paragraph{Answer-side statistics remain stronger for correctness selection.} Given a
same-question cohort, $\deff$ predicts majority-vote correctness at AUC $0.506$
versus $0.832$ for a vote margin, and same-answer-support coherence is
near-identical for correct and incorrect answers ($0.698$ vs.\ $0.697$; AUC
$0.483$). (These are different estimands, not two conventions for one number:
\Cref{tab:app:correctness} scores the \emph{signed} terminal-frame
($\bbudget{=}32768$) $\log\deff$ of the full boxed cohort graph on the $\Naudit$
cohorts with $\ge10$ answer-assigned rollouts, while this paragraph scores a
\emph{direction-folded} $\deff$ recomputed on the answer-bearing subgraph at the
$\bbudget{=}4096$ prefix over $3{,}055$ cohorts with $\ge16$ answer-bearing
rollouts---folding is why this value cannot fall below $0.5$. Neither analysis
finds any routing feature competitive with the answer-side baselines, and $r$'s
resolved below-chance value is not inverted into an error signal.)
The pooled $0.506$ should not be read as ``the signal is absent'': resolved by
granularity, routing reaches $0.65$ across problems and $0.68$ within a problem,
while each remains below a matched text statistic on its own terms (problem prior
$0.95$, answer agreement $0.75$, vote margin $0.83$), and the single-rollout null
is powered (AUC $0.51$, power $1.00$ against a $5$-point effect) rather than
data-limited. The comparison holds at every granularity, separating answer selection
from structural monitoring.
This is structural: the routing graph encodes graded, diffuse cohort organization
rather than a \emph{label-free}
answer-separating partition---even the oracle commit graph has no hard clusters
(eigengap $\hat k\!\approx\!1$, spectral clustering~\citep{vonluxburg2007tutorial}
reaching an adjusted Rand index~\citep{hubert1985comparing} of only
${\approx}0.07$).

\paragraph{Prefix state is descriptive rather than prognostic.} Prefix routing state does
not predict whether a cohort will \emph{ultimately reach unanimity} (a GRPO
zero-advantage proxy): pooled AUC is only $0.45$--$0.56$. The state describes
present organization, not destination.

\paragraph{Hard answer labels subsume the tested soft-geometry weighting.} With
$m$ hard labels
in hand, the tested soft-geometry weightings are subsumed:
graph-weighted distribution extrapolation matches a plain estimate
($\mathrm{TV}$ $0.0980$ vs.\ $0.0983$), and density-weighted voting performs
comparably to
equal-weight voting (majority-vote accuracy $0.9496$ vs.\ $0.9512$ on this
control's evaluation base, which differs from the frontier base of
\Cref{sec:app:frontier}). The graph's value is concentrated \emph{before} labels
are available---in geometric calibration.

\paragraph{Oracle-seeded propagation remains limited by diffuse membership.}
Factorizing a
seed-and-propagate pipeline for majority-vs-rest separation shows that the
limiting factor is
membership assignment rather than seed coverage: selectors cover minority members
($0.30$--$0.45$ minority share), but $16$-seed subgraph purity is only
$0.55$--$0.68$, and even \emph{ground-truth} seed labels cap propagation at $0.616$
(prefix) / $0.692$ (commit). The full feasible pipeline sits at chance
($0.48$--$0.51$).

\paragraph{$\deff$ is better suited to observation than vote reweighting.} We
evaluated six ways of incorporating $\deff$ into
voting---re-weighting, stop trigger, vote count, subset selection, confidence, and
efficiency---and none improves on the corresponding answer-side baseline under
the tested protocols. Diversity/effective-count re-weighting (using
$\deff(\Wmat)$ as an effective vote count) slightly \emph{hurts} majority vote at
commit ($-0.0072$, significant): the re-weighting quantity carries no correctness
signal (the coherence null above). An out-of-fold logistic over vote margin and gap
(on the same $3{,}055$-cohort pool, $0.833$) gains nothing from adding $\deff$ and
$\dens$ ($0.830$).

\subsection{Answer/Compute Frontier Analysis}
\label{sec:app:frontier}

We also tested directly whether these routing summaries extend the answer/compute
frontier; under the tested protocols the strongest frontier points come from
answer-side signals, cleanly separating the structural-monitoring role of routing
spectra from answer-side selection and stopping.

\paragraph{Token$\times$accuracy Pareto.} On $2{,}996$ problems with exact
per-rollout lengths, the strongest frontier points are provided by the number of
votes and text-margin
early-stopping: plain self-consistency peaks at $N{\approx}24$ ($0.7613$ at
$40.9\%$ cost, above full $N{=}64$'s $0.7570$---over-provisioning hurts),
consistency early-stop is an on-frontier saver, and the tested routing-based
levers
(redundancy kill, $-0.3$ to $-0.6$\,pp; $\deff$-based cross-problem $N$
allocation) do not reach the frontier.

\paragraph{Early-stop discrimination.} When consistency early-stop fires ($62\%$ of
problems, premature-stop rate $0.001$), predicting whether a stop is correct is AUC
$0.660$ for the vote margin versus $0.521$ for $\deff$; routing adds $-0.034$
$[-0.076,+0.025]$ over the margin (CI includes zero).

\paragraph{Vote confidence and efficiency.} Confidence and efficiency are likewise
the margin's doing: $92.4\%$ of problems are settled by $N{=}8$, and truncating the
most-confident $70\%$ at $N{=}8$ reaches full-$64$ accuracy at $24.8$ votes.
Routing provides no measurable incremental gain in these answer-side decisions.

\section{Reproducibility and Sample Accounting}
\label{sec:app:repro}

All cohort statistics derive from one fixed activation cache of MoE routing
histograms, with $\deff$ computed by the same symmetrize-and-clip implementation
throughout; the commit graph uses the boxed-answer subwindow. The corpus is
$\Ncohorts$ cohorts across $50$ model$\times$dataset shards, identical under the
weighted and the indicator (binary) graph; trajectory statistics, including the
$98.5\%$ prominent-peak rate, are computed on the weighted graph over all
$\Ncohorts$ cohorts, and $\Naudit$ is the audit family's retention ($\ge10$ graded
rollouts), not a pipeline difference. The spectral-anatomy analyses
(\Cref{sec:app:anatomy}) evaluate \Cref{eq:decomp} on the $\Nident$ frames with
defined $\dperp$ (of $\Nframes$ total); the $\Ncomplete$ complete cells supply the
attribution segments and the $\Ncvframes$ CV-fit frames. The audit ladders
(\Cref{sec:app:audit}) are anchored to the corpus tables to
deviation $<10^{-13}$. Every analysis draws from this one cache; the retained count
differs only by an explicit per-analysis filter (\Cref{tab:app:accounting}), never
by a dataset swap.

\paragraph{Infrastructure and availability.}
All rollouts were generated with vLLM on a single node with eight NVIDIA H20 GPUs
and $2$\,TB of system memory; the routing histograms were captured during serving,
and all downstream spectral and statistical analyses ran on the fixed activation
cache described above. The activation caches, cohort-level tables, and the complete
analysis code will be released under a license permitting free research use, with
exact software versions pinned in the release.

\end{document}